%% file: main.tex
\RequirePackage[dvipsnames,table]{xcolor} 
\documentclass[10pt,twocolumn,letterpaper]{article}

\usepackage[accsupp]{axessibility}
\usepackage{iccv}
\usepackage{times}
\usepackage{epsfig}
\usepackage{graphicx}
\usepackage{amsmath}
\usepackage{amssymb}
\usepackage{enumitem}

\usepackage{balance}
\usepackage{comment}
\usepackage{booktabs}
\usepackage{caption}
\usepackage{subcaption}
\usepackage{pifont}
\newcommand{\p}[1]{\vspace{1mm}\noindent\textbf{#1}}

\newcommand{\vilbert}{ViLBERT}
\newcommand{\vlnbert}{VLN-BERT}
\newcommand{\airbert}{Airbert}
\newcommand{\airbnb}{BnB}
\newcommand{\mask}{\texttt{[MASK]}}
\newcommand{\cls}{\texttt{[CLS]}}
\newcommand{\sep}{\texttt{[SEP]}}
\newcommand{\imgtok}{\texttt{[IMG]}}
\newcommand{\mcV}{\mathcal{V}}
\newcommand{\mcN}{\mathcal{N}}

\usepackage{multirow}

\definecolor{Gray}{gray}{0.87}
\newcolumntype{h}{>{\columncolor{Gray}}c}
\newcommand{\h}{\cellcolor{Gray}}
\newcommand{\start}{\ding{108}}
\newcommand{\final}{\ding{110}}
\newcommand{\success}{{\color{ForestGreen}{\ding{52}}}}
\newcommand{\failure}{{\color{Bittersweet}{\ding{56}}}}

\newcommand*{\email}{\includegraphics[scale=0.09]{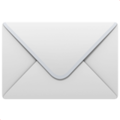}}%
\newcommand*{\website}{\includegraphics[scale=0.07]{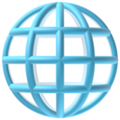}}%

\usepackage[pagebackref=true,breaklinks=true,colorlinks,bookmarks=false]{hyperref}

\iccvfinalcopy %

\def\iccvPaperID{2609} %

\ificcvfinal\fi

\begin{document}

\title{Airbert: In-domain Pretraining for Vision-and-Language Navigation}

\author{Pierre-Louis Guhur$^{1}$,
Makarand Tapaswi$^{2}$,
Shizhe Chen$^{1}$,
Ivan Laptev$^{1}$,
Cordelia Schmid$^{1}$\\
$^{1}$Inria, \'Ecole normale sup\'erieure, CNRS, PSL Research University, Paris, France\\
$^{2}$IIIT Hyderabad, India\\
{\tt\small  \email~pierre-louis.guhur@inria.fr}  ~~
{\tt\small \website~\url{https://airbert-vln.github.io}}
}

\maketitle
\ificcvfinal\thispagestyle{empty}\fi

\input{sections/0.abstract.tex}

\input{sections/1.intro.tex}
\input{sections/2.related-work.tex}

\input{sections/3.airbnb-dataset.tex}
\input{sections/4.airbert.tex}
\input{sections/6.experiments.tex}
\input{sections/7.conclusion.tex}

{\small \noindent\textbf{Acknowledgments.}
This work was granted access to the HPC resources of IDRIS under the allocation 101002 made by GENCI. 
It was funded in part by the French government under management of Agence Nationale de la Recherche as part of the ''Investissements d'avenir'' program, reference ANR19-P3IA-0001 (PRAIRIE 3IA Institute) and by Louis Vuitton ENS Chair on Artificial Intelligence.
}

\balance
{\small
\bibliographystyle{ieee_fullname}
\bibliography{biblio/iccv21, biblio/shortstrings}
}

\newpage
\appendix
\section*{Appendix}
\input{sections/supp.tex}

\end{document}

%% file: sections/0.abstract.tex
\begin{abstract}
Vision-and-language navigation (VLN) aims to enable embodied agents to navigate in realistic environments using natural language instructions.
Given the scarcity of domain-specific training data and the high diversity of image and language inputs, the generalization of VLN agents to unseen environments remains challenging.
Recent methods explore pretraining to improve generalization, however, the use of generic image-caption datasets 
or existing small-scale VLN environments 
is suboptimal and results in limited improvements.
In this work, we introduce \airbnb\footnote{Bed and Breakfast}, a large-scale and diverse in-domain VLN dataset. 
We first collect image-caption (IC) pairs from hundreds of thousands of listings from online rental marketplaces.
Using IC pairs we next propose automatic strategies to generate millions of VLN path-instruction (PI) pairs.
We further propose a shuffling loss that improves the learning of temporal order inside PI pairs.
We use \airbnb{} to pretrain our \airbert\footnote{
\emph{\airbert} is an Old Irish word meaning \emph{practice}, here referring to model pretraining on pretext tasks similar to VLN.}
model that can be adapted to discriminative and generative settings and show that it outperforms state of the art for Room-to-Room (R2R) navigation and Remote Referring Expression (REVERIE) benchmarks.
Moreover, our in-domain pretraining significantly increases performance on a challenging few-shot VLN evaluation, where we train the model only on VLN instructions from a few houses.
\end{abstract}

%% file: sections/1.intro.tex
\section{Introduction}
\label{sec:intro}

\begin{figure}[t]
\centering
\includegraphics[width=\columnwidth]{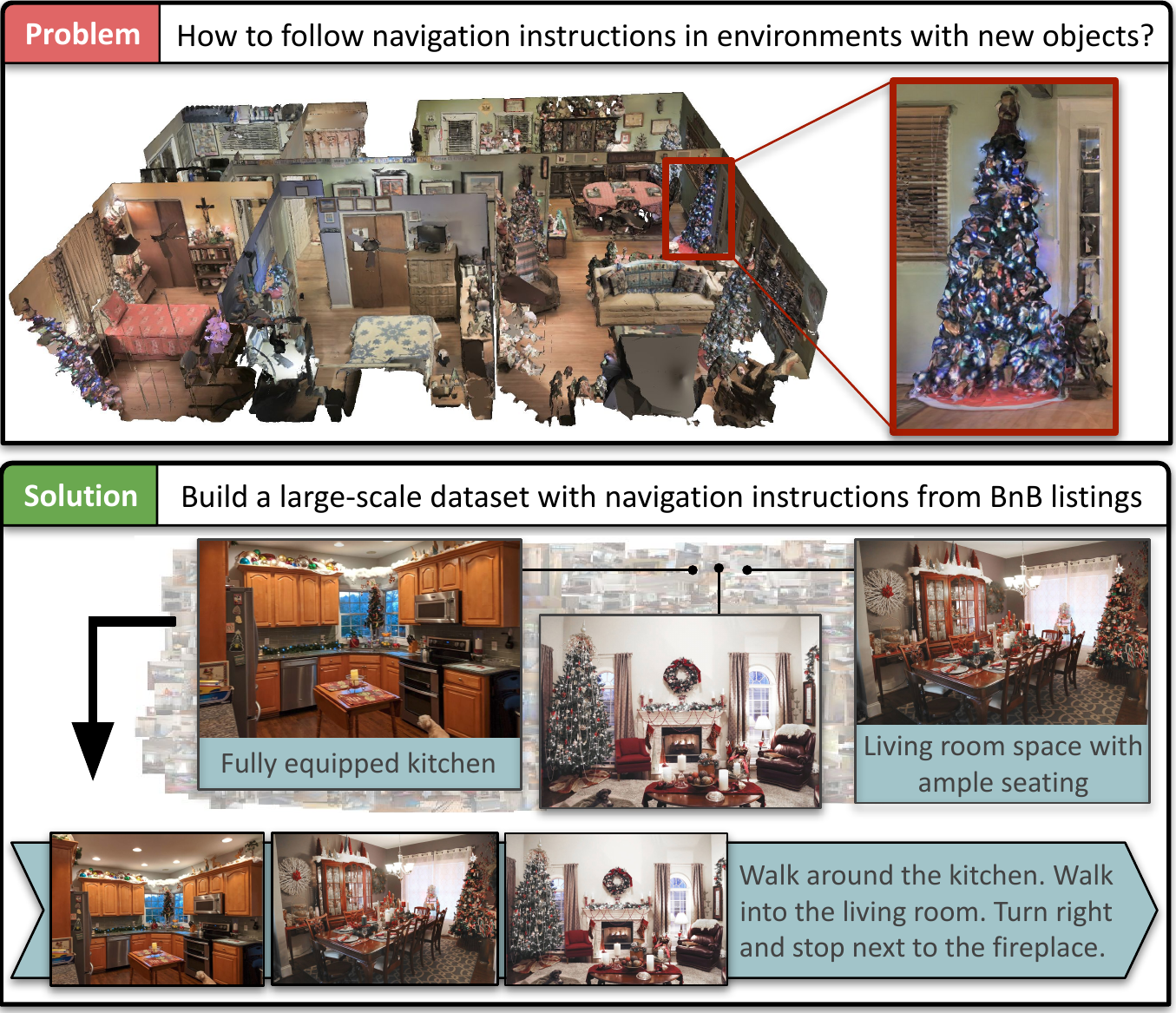}
\vspace{-6mm}
\caption{VLN tasks are evaluated on unseen environments at test time.
\emph{Top}: None of the training houses contain a Christmas theme making this test environment particularly challenging.
\emph{Bottom}: We build a large-scale, visually diverse, and in-domain dataset by creating path-instruction pairs close to a VLN-like setup and show the benefits of self-supervised pretraining.}
\vspace{-5mm}
\label{fig:teaser}
\end{figure}

In vision-and-language navigation (VLN), an agent is asked to navigate in home environments following natural language instructions~\cite{anderson2018evaluation,anderson2018r2r}.
This task is attractive to many real-world applications such as domestic robotics and personal assistants.
However, given the high diversity of VLN data across environments and the difficulty of the manual collection and annotation of VLN training data at scale, the performance of current methods remains limited, especially for previously unseen environments~\cite{zhangdiagnosing}.

Our work is motivated by significant improvements in vision and language pretraining~\cite{alberti2019b2t2,chen2020uniter,li2020oscar,lu2019vilbert,lu2020_12in1,su2019vlbert}, where deep transformer models~\cite{vaswani2017attention} are trained via self-supervised proxy tasks~\cite{devlin2018bert} using large-scale, automatically harvested image-text datasets~\cite{ordonez2011sbu,ConceptualCaptions}.
Such pretraining enables learning transferable multi-modal representations achieving state-of-the-art performance in various vision and language tasks.
Similarly, with the goal of learning an embodied agent that generalizes, recent works~\cite{hao2020prevalent,huang2019transferable,li2019press,majumdar2020vlnbert} have explored different pretraining approaches for VLN tasks.

In~\cite{hao2020prevalent,huang2019transferable}, annotated path-instruction pairs
are augmented with a \emph{speaker} model that generates instructions for random unseen paths.
However, as these paths originate from a small set of 61 houses used during training, they are limited in visual diversity.
The limited pretraining environments do not equip agents with visual understanding abilities that enable generalization to unseen houses, see Fig.~\ref{fig:teaser}.
To address this problem, \vlnbert~\cite{majumdar2020vlnbert} proposes to pretrain the agent on generic image-caption datasets that are abundant and cover diverse visio-linguistic knowledge.
However, these image-caption pairs are quite different from the dynamic visual stream (path) and navigable instructions observed by a VLN agent.
Such out-of-domain pretraining, although promising, only brings limited gains to the navigation performance.
Besides the above limitations, existing pretraining methods do not place much emphasis on temporal reasoning abilities in their proxy tasks such as one-step action prediction~\cite{hao2020prevalent} and path-instruction pairing~\cite{majumdar2020vlnbert}, while such reasoning is important to a sequential decision making task like VLN.
As a result, even if performance in downstream tasks is improved, the pretrained models may still be brittle.
For example, a simple corruption of instructions by swapping noun phrases within the instruction, or replacing them with other nouns, leads to significant confusion as models are unable to pick the correct original pair. 

In this paper, we explore a different data source and proxy tasks to address the above limitations in pretraining a generic VLN agent.
Though navigation instructions are rarely found on the Internet, image-caption pairs from home environments are abundant in online marketplaces (\eg~\emph{Airbnb}), which include images and descriptions of rental listings.
We collect \airbnb, a new large-scale dataset with 1.4M indoor images and 0.7M captions.
First, we show that in-domain image-caption pairs bring additional benefits for downstream VLN tasks when applied with generic web data~\cite{majumdar2020vlnbert}.
In order to further reduce the domain gap between the \airbnb~pretraining and the VLN task, we present an approach to transform static image-caption pairs into visual paths and navigation-like instructions (Fig.~\ref{fig:teaser} bottom), leading to large additional performance gains.
We also propose a shuffling loss that improves the model's temporal reasoning abilities by learning a temporal alignment between a path and the corresponding instruction.

Our pretrained model, \airbert, is a generic transformer backbone that can be readily integrated in both discriminative VLN tasks such as path-instruction compatibility prediction~\cite{majumdar2020vlnbert} and generative VLN tasks~\cite{hong2021recurrentvln} in R2R navigation~\cite{anderson2018r2r} and REVERIE remote referring expression~\cite{qi2020reverie}.
We achieve state-of-the-art performance on these VLN tasks with our pretrained model.
Beyond the standard evaluation, our in-domain pretraining opens an exciting new direction of \emph{one/few-shot VLN} where the agent is trained on examples only from one/few environment(s) and expected to generalize to other unseen environments.

In summary, the contributions of this work are three-fold.
(1) We collect a new large-scale in-domain dataset, \airbnb, to promote pretraining for vision-and-language navigation tasks. 
(2) We curate the dataset in different ways to reduce the distribution shift between pretraining and VLN and also propose the shuffling loss to improve temporal reasoning abilities.
(3) Our pretrained \airbert~can be plugged into generative or discriminative architectures and achieves state-of-the-art performance on R2R and REVERIE datasets.
Moreover, our model generalizes well under a challenging one/few-shot VLN evaluation, truly highlighting the capabilities of our learning paradigm.

%% file: sections/2.related-work.tex
\section{Related work}
\label{sec:related}

\p{Vision-and-language navigation.}
VLN~\cite{anderson2018r2r} has received significant attention with a large number of followup tasks introduced in recent years~\cite{anderson2018evaluation, chen2019touchdown, krantz2020r2rce, ku2020rxr, nguyen2019hanna, nguyen2019vlna, qi2020reverie, shridhar2020alfred, thomason2020cvdn}. 
Early days of VLN saw the use of sequence-to-sequence LSTMs to predict low-level actions~\cite{anderson2018r2r} or high-level directions in a panoramic action space~\cite{fried2018speaker}.
Different attention mechanisms \cite{ma2019self,qi2020object} are proposed to improve cross-modal alignment.
Various reinforcement learning based training algorithms \cite{tan2019envdrop,wang2020serl,wang2019reinforced,wang2018look} and searching algorithms in inference \cite{fried2018speaker,ma2019self,ma2019regretful} have also been explored to improve the VLN performance.

To improve an agent's generalization to unseen environments, data augmentation is performed by using a \emph{speaker} model~\cite{fried2018speaker} that generates instructions for random paths in seen environments, and environment dropout~\cite{tan2019envdrop} is used to mimic new environments.
While pretraining LSTMs for transferable representations is adopted by~\cite{huang2019transferable}, recently, there has been a shift towards transformer models~\cite{hao2020prevalent} to learn generic multimodal representations.
This is further extended to a recurrent model that significantly improves sequential action prediction~\cite{hong2021recurrentvln}.
However, the limited environments in pretraining~\cite{hao2020prevalent,huang2019transferable} constrain the generalization ability to unseen scenarios.
Most related to this work, \vlnbert~\cite{majumdar2020vlnbert} transfers knowledge from abundant, but out-of-domain image-text data to improve path-instruction matching.
In this work, we not only create a large-scale, \emph{in-domain} \airbnb~dataset, but also propose effective pretraining strategies to mitigate the domain-shift between webly crawled image-text pairs and VLN data.

\p{Large-scale visio-linguistic pretraining.}
Thanks to large-scale image-caption pairs automatically collected from the web~\cite{miech2019howto100m, ordonez2011sbu, radford2021learning, ConceptualCaptions}, visio-linguistic pretraining (VLP) has made great breakthroughs in recent years.
Several VLP models~\cite{chen2020uniter, li2020oscar, lu2019vilbert, tan2019lxmert} have been proposed based on the transformer architecture~\cite{vaswani2017attention}.
These models are often pretrained with self-supervised objectives akin to those in BERT~\cite{devlin2018bert}:
masked language modeling, masked region modeling and vision-text pairing.
Fine-tuning them on downstream datasets achieves state-of-the-art performance on various VL tasks~\cite{antol2015vqa, kazemzadeh2014referitgame, wang2016learning, vinyals2016show}.
While such pretraining focuses on learning correlations between vision and text, it is not designed for sequential decision making as required in embodied VLN.
The goal of this work is not to improve VLP architectures but to present in-domain training strategies that lead to performance improvements for VLN tasks.

%% file: sections/3.airbnb-dataset.tex
\section{\airbnb~Dataset}
\label{sec:airbnb}

Hosts that rent places on online marketplaces often upload attractive and unique photos along with descriptions.
One such marketplace, \emph{Airbnb}, has 5.6M listings from over 100K cities all around the world~\cite{airbnb}.
We propose to use this abundant and curated data for large-scale in-domain VLN pretraining.
In this section, we first describe how we collect image-caption  pairs from \emph{Airbnb}.
Then, we propose methods to transform images and captions into VLN-like path-instruction pairs to reduce the domain gap between webly crawled image-caption pairs and VLN tasks (see Fig.~\ref{fig:dataset}).

\subsection{Collecting \airbnb~Image-Caption Pairs}
\p{Collection process.}
We restrict our dataset to listings from the US (about 10\% of \emph{Airbnb}) to ensure high quality English captions and visual similarity with Matterport environments~\cite{Matterport3D}.
The data collection proceeds as follows: 
(1)~obtain a list of locations from Wikipedia; 
(2)~find listings in these locations by querying the \emph{Airbnb} search engine; 
(3)~download listings and their metadata; 
(4)~remove \emph{outdoors} images\footnote{While outdoor images may contain interesting features (\eg~a patio), we observe that removing them increases performance.} as classified by a ResNet model pretrained on Places365~\cite{zhou2017places}; and
(5)~remove invalid image captions such as emails, URLs and duplicates.

\p{Statistics.}
We downloaded almost 150k listings and their metadata (1/4 of the listings in the US) in step 3, leading to over 3M images and 1M captions.
After data cleaning with steps 4 and 5, we obtain 713K image-caption pairs and 676K images without captions.
Table~\ref{tab:bnb_dataset_cmpr} compares our \airbnb~dataset to other datasets used in previous works for VLN (pre-)training.
It is larger than R2R~\cite{anderson2018r2r}, REVERIE~\cite{qi2020reverie} and includes a large diversity of rooms and objects, which is not the case for Conceptual Captions~\cite{ConceptualCaptions}. 
We posit that such in-domain data is crucial to deal with the data scarcity challenge in VLN environments as illustrated in Fig.~\ref{fig:teaser}.
We use 95\% of our \airbnb~dataset for training and the remaining 5\% for validation.

Apart from images and captions, our collected listings contain structured data including a list of amenities, a general description, reviews, location, and rental price, which may offer additional applications in the future.
More details about the dataset and examples are presented in the Appendix~\ref{sec:supp-bnb}. 

\input{tables/bnb_dataset}

\subsection{Creating \airbnb~Path-Instruction Pairs}
\label{sec:create_pi_pairs}
\airbnb~image-caption (IC) pairs are complementary to Conceptual Captions (ConCaps) as they capture diverse VLN environments.
However, they still have large differences from path-instruction (PI) pairs in VLN tasks.
For example, during navigation, an agent observes a sequence of panoramic views rather than a single image, and the instruction may contain multiple sentences.
To mitigate this domain gap, we propose strategies to automatically craft path-instruction pairs starting from \airbnb-IC pairs.

\subsubsection{Concatenating Images and Texts in a \airbnb~Listing}
\label{sec:concate_method}
Images in a \airbnb~listing usually depict different locations in a house, mimicking the sequential visual observations an agent makes while navigating in the house.
To create a VLN-like path-instruction pair,  we randomly select and concatenate $K$\footnote{typically 4 - 7 to match the number of steps in the R2R dataset} image-caption pairs from the listing.
In between each caption, we randomly add a word from ``\emph{and}'', ``\emph{then}'', ``.'' or nothing to make the concatenated instruction more fluent and diverse.

\subsubsection{Augmenting \emph{Paths} with Visual Contexts}
In the above concatenated path, each location only contains one \airbnb~image, and perhaps with a limited view angle as hosts may focus on objects or amenities they wish to highlight.
Therefore, it lacks the panoramic visual context at each location that the agent receives in real navigation paths.
Moreover, each location in the concatenated instruction is described by a unique sentence, while adjacent locations are often expressed together in one sentence in VLN instructions~\cite{hong2020fgr2r}.
To address the above issues with concatenation, we propose two approaches to compose paths that have more visual context and can also leverage the abundant images without captions (denoted as \emph{captionless images}).

\begin{figure}[t]
\centering
\includegraphics[width=\columnwidth]{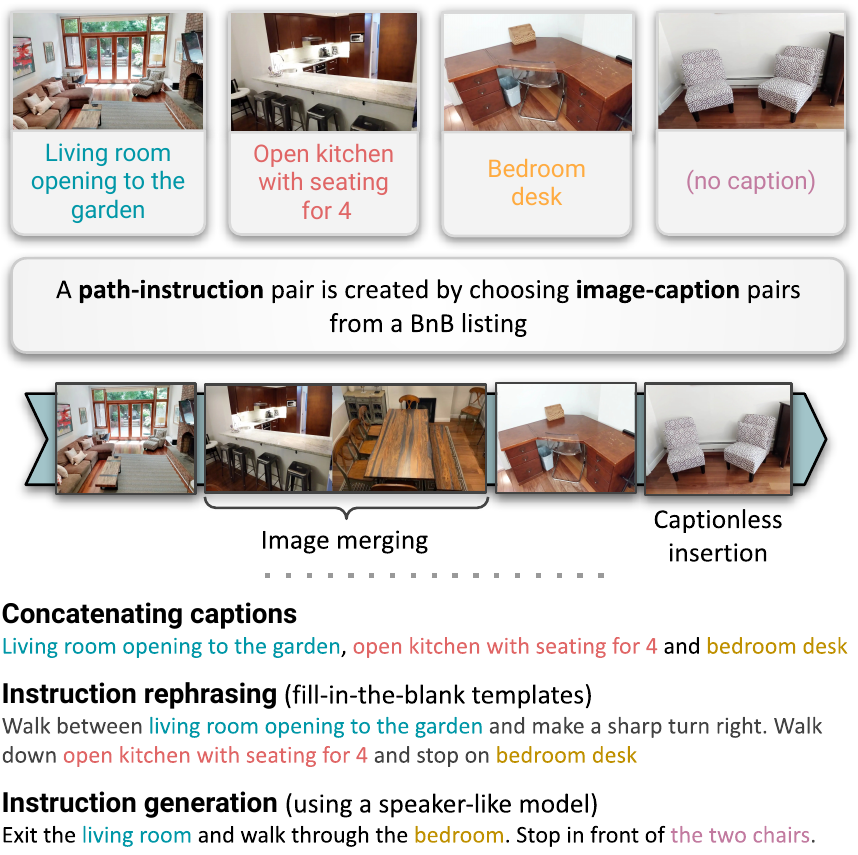}
\caption{We explore several strategies to automatically create navigation-like instructions from image-caption pairs.
}
\label{fig:dataset}
\end{figure}

\p{1. Image merging}
extends the panoramic context of a location by grouping images from similar room categories (see Fig.~\ref{fig:dataset}).
For example, if the image depicts a kitchen sink, it is natural to expect images of other objects such as forks and knives nearby.
Specifically, we first cluster images of similar categories (\eg~\emph{kitchen}) using room labels predicted by a pretrained Places365 model~\cite{zhou2017places}.
Then, we extract multiple regions from this \emph{merged} set of images, and use them as an approximation to the panoramic visual representation.


\p{2. Captionless image insertion.}
Table 1 shows that half of the BnB images are captionless. Using them allows to increase the size of the dataset. When creating a path-instruction pair from the concatenation approach, a captionless image is inserted as if its caption was an empty string.
The BnB PI pairs hence better approximate the distribution of the R2R path-instructions: (1) some images in the path are not described
and (2)~instructions have similar number of noun phrases.

\subsubsection{Crafting \emph{Instructions} with Fluent Transitions}
The concatenated captions mainly describe rooms or objects at different locations, but do not contain any of the actionable verbs as in navigation instructions, \eg~``\emph{turn left at the door}'' or ``\emph{walk straight down the corridor}''.
We suggest two strategies to create fake instructions that have fluent transitions between sentences.

\p{1. Instruction rephrasing.}
We use a fill-in-the-blanks approach to replace noun-phrases in human annotated navigation instructions~\cite{anderson2018r2r} by those in \airbnb~captions (see Fig.~\ref{fig:dataset}).
Concretely, we create more than 10K instruction templates containing 2-7 blanks, and fill the blanks with noun-phrases extracted from \airbnb~captions.
The noun-phrases matched to object categories from the Visual Genome~\cite{krishna2017vg} dataset are preferred during selection.
This allows us to create VLN-like instructions with actionable verbs interspersed with room and object references for visual cues that are part of the \airbnb~path (see Fig.~\ref{fig:dataset}).



\p{2. Instruction generation} is a video captioning like model that takes in a sequence of images and generates an instruction corresponding to an agent's path through an environment.
To train this model, we adopt ViLBERT and train it to generate captions for single BnB image-caption pairs.
Further, this model is fine-tuned on trajectories of the R2R dataset to generate instructions.
Finally, we use this model to generate BnB PI pairs by producing an instruction for a concatenated image sequence from BnB (the path).



%% file: tables/bnb_dataset.tex
\begin{table}[t]
\centering
\tabcolsep=0.11cm
\small
\begin{tabular}{ll ccc}
\toprule
Dataset & Source & \#Envs & \#Imgs & \#Texts \\
\midrule
R2R~\cite{anderson2018r2r} & Matterport    & 90 & 10.8K & 21.7K \\
REVERIE~\cite{qi2020reverie} & Matterport  & 86 & 10.6K & 10.6K \\
Speaker~\cite{tan2019envdrop} & Matterport & 60 & 7.8K & 0.2M \\
\midrule
ConCaps~\cite{ConceptualCaptions} & Web images & - & 3.3M & 3.3M \\
\textbf{\airbnb} (ours) & Airbnb & 140K & 1.4M & 0.7M \\
\bottomrule
\end{tabular}
\vspace{-2mm}
\caption{Comparing \airbnb~to other existing VLN datasets.
The \#images from Matterport environments~\cite{Matterport3D} refers to the \#panoramas.
The speaker model~\cite{tan2019envdrop} generates instructions for randomly selected trajectories, but is limited to panoramas from 60 training environments.
Note that the data from Conceptual Captions (ConCaps) may feature some houses, but it is not the main category.
}
\vspace{-4mm}
\label{tab:bnb_dataset_cmpr}
\end{table}

%% file: sections/4.airbert.tex
\section{\airbert: A Pretrained VLN Model}
\label{sec:airbert}

\begin{figure*}
\centering
\includegraphics[width=\linewidth]{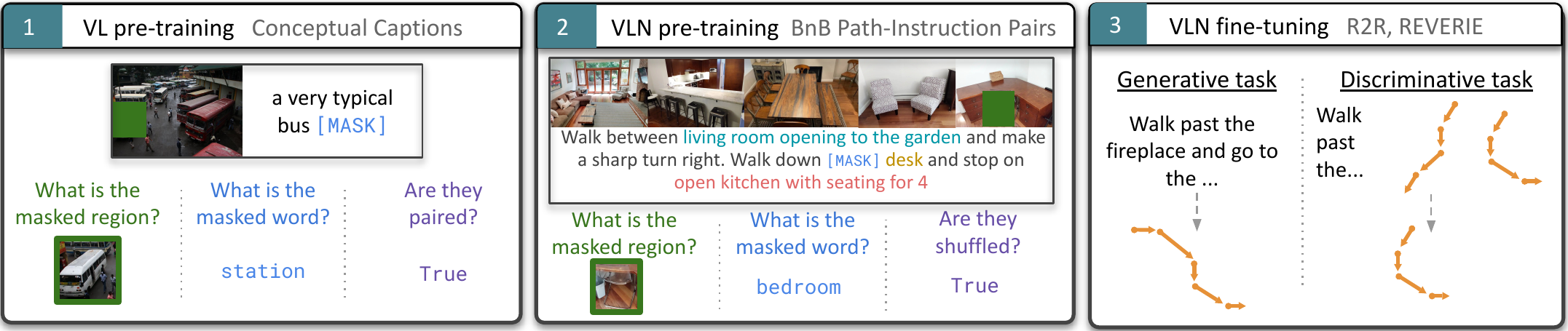}
\caption{Overview of our pretraining approach.
Instead of the usual VL pretraining (panel 1), we adopt in-domain data and use the path-instruction pairs to train \airbert~with the masking and shuffling losses (panel 2).
We fine-tune Airbert on downstream VLN tasks using both discriminative or generative models (panel 3).
}
\vspace{1mm}
\label{fig:model}
\end{figure*}

In this section, we present \airbert, our multi-modal transformer pretrained on the \airbnb~dataset with masking and shuffling losses.
We first introduce the architecture of~\airbert, and then describe datasets and pretext tasks in pretraining.
Finally, we show how \airbert~can be adapted to downstream VLN tasks.

\subsection{\vilbert-like Architecture}
\vilbert~\cite{lu2019vilbert} is a multi-modal transformer extended from BERT~\cite{devlin2018bert} to learn joint visio-linguistic representations from image-caption pairs, as illustrated in Fig.~\ref{fig:model}.

Given an image-caption pair $(V, C)$, the model encodes the image as region features $[v_1, \ldots, v_\mcV ]$ via a pretrained Faster R-CNN~\cite{anderson2017butd}, and embeds the text as a series of tokens: $[\cls, w_1, \ldots, w_T, \sep]$, where \cls and \sep are special tokens added to the text.
\vilbert~contains two separate transformers that encode $V$ and $C$ and it learns cross-modal interactions via co-attention~\cite{lu2019vilbert}.

We follow a similar strategy to encode path-instruction pairs (created in Sec.~\ref{sec:create_pi_pairs}) that contain multiple images and captions $\{(V_k, C_k)\}_{k=1}^K$.
Here, each $V_k$ is represented as visual regions $v^k_i$ and $C_k$ as word tokens $w^k_t$.
Respectively, the visual and text inputs to \airbert~are:
{\small
\begin{align}
X_V &= [\imgtok, v^1_1, \ldots, v^1_{\mcV_1}, \ldots, \imgtok, v^K_1, \ldots, v^K_{\mcV_K}], \\
X_C &= [\cls, w^1_1, \ldots, w^1_{T_1}, \ldots, w^K_1, \ldots, w^K_{T_K}, \sep ] ,
\end{align}}%
where the \imgtok~token is used to separate image region features taken at different locations.

Note that while our approach is not limited to a \vilbert-like architecture, we choose \vilbert~for a fair comparison with previous work~\cite{majumdar2020vlnbert}.

\subsection{Datasets and Pretext Tasks for Pretraining}
We use Conceptual Captions (ConCaps)~\cite{ConceptualCaptions} and \airbnb-PI in subsequent pretraining steps (see Fig.~\ref{fig:model}) to reduce the domain gap for downstream VLN tasks.

Previous multi-modal pretraining efforts~\cite{lu2019vilbert, majumdar2020vlnbert, huang2019transferable} commonly use two self-supervised losses given image-caption~(IC) pairs or path-instruction (PI) pairs:
(1) \emph{Masking} loss:
An input image region or word is randomly replaced by a \mask~token.
The output feature of this masked token is trained to predict the region label or the word given its multi-modal context.
(2) \emph{Pairing} loss: 
Given the output features of \imgtok and \cls~tokens, a binary classifier is trained to predict whether the image (path) and caption (instruction) are paired.

The above two pretext tasks mainly focus on learning object-word associations instead of reasoning about the temporal order of paths and instructions.
For example, if an image $V_i$ appears before $V_j$, then words from its caption $C_i$ should appear before $C_j$.
In order to promote such a temporal reasoning ability, we propose an additional \emph{shuffling} loss to enforce alignment between PI pairs.

Given an aligned PI pair $X^+ = \{(V_k, C_k)\}_{k=1}^K$, we generate $\mcN$ negative pairs $X^-_n = \{(V_k, C_l)\}, k\neq l$, by shuffling the composed images or the captions.
We train our model to choose the aligned PI pair as compared to the shuffled negatives by minimizing the cross-entropy loss:
\begin{equation}
L = -\log \frac{\exp(f(X^+))}{\exp(f(X^+)) + \sum_n \exp(f(X^-_n))} \, ,
\end{equation}
where $f(X)$ denotes the similarity score (logit) computed via \airbert~for the PI pair $X$.

\subsection{Adaptations for Downstream VLN tasks}
\label{sec:downstream}
We consider two VLN tasks: goal-oriented navigation (R2R \cite{anderson2018r2r}) and object-oriented navigation (REVERIE \cite{qi2020reverie}).
\airbert~can be readily integrated in discriminative and generative models for the above VLN tasks.

\p{\mbox{Discriminative Model: Navigation as Path-Selection~\cite{majumdar2020vlnbert}.}}
The navigation problem on the R2R dataset is formulated as a path selection task in~\cite{majumdar2020vlnbert}.
Several candidate paths are generated via beam search from a navigation agent such as~\cite{tan2019envdrop}, and a discriminative model is trained to choose the best path among them.
We fine-tune \airbert~on the R2R dataset for path selection.
A two-stage fine-tuning process is adopted: 
in the first phase, we use \emph{masking} and \emph{shuffling} losses on the PI pairs of the target VLN dataset in a manner similar to \airbnb~PI pairs;
in the second phase, we choose a positive candidate path as one that arrives within 3m of the goal, and contrast it against 3 negative candidate paths.
We also compare multiple strategies to mine additional negative pairs (other than the 3 negative candidates), and in fact, empirically show that negatives created using shuffling outperform other options.

\p{Generative Model: Recurrent VLN-BERT~\cite{hong2021recurrentvln}.}
The Recurrent VLN-BERT model adds recurrence to a state in the transformer to sequentially predict actions, achieving state-of-the-art performance on R2R and REVERIE tasks.
We use our \airbert~architecture as its backbone and apply it to the two tasks as follows.
First, the language transformer encodes the instruction via self-attention.
Then, the embedded \cls~token in the instruction is used to track history and concatenated with visual tokens (observable navigable views or objects) in each action step.
Self-attention and cross-attention on embedded instructions are employed to update the state and visual tokens and the attention score from the state token to visual tokens is used to decide the action at each step.
We fine-tune the Recurrent VLN-BERT model with \airbert~as the backbone in the same way as~\cite{hong2021recurrentvln}.

Additional details about the models and their implementation are provided in the Appendix~\ref{sec:supp-details}.


%% file: sections/6.experiments.tex
\section{Experimental Results}
\label{sec:xp}

We first perform ablation studies evaluating alternative ways to pretrain \airbert~in Sec.~\ref{sec:xp_pretrain_airbert}.
Then, we compare \airbert~with state-of-the-art methods on R2R and REVERIE tasks in Sec.~\ref{sec:xp_sota}.
Finally, in Sec.~\ref{sec:eval:fsl}, we evaluate models in a more challenging setup: VLN few-shot learning where an agent is trained on examples taken from one/few houses.



\input{tables/how.tex}
\input{tables/shuffle.tex}

\input{tables/dataaug-r2r.tex}
\p{R2R Setup.}
Most of our experiments are conducted on the R2R dataset~\cite{anderson2018r2r}, where we adopt standard splits and metrics defined by the task.
We focus on success rate (SR), which is the ratio of predicted paths that stop within 3m of the goal.
Please refer to~\cite{anderson2018r2r, majumdar2020vlnbert} for a more detailed explanation of the metrics.
In particular, as the discriminative model uses path selection for R2R, we follow the pre-explored environment setting adopted by \vlnbert~\cite{majumdar2020vlnbert}.


\p{REVERIE Setup.}
We also adopt standard splits and metrics on the REVERIE task~\cite{qi2020reverie}.
Here, the success rate (SR) is the ratio of paths for which the agent stops at a viewpoint where the target object is visible.
Remote Grounding Success Rate (RGS) measures accuracy of localizing the target object in the stopped viewpoint, and RGS per path length (RGSPL) is a path length weighted version.


\input{tables/analysis.tex}

\input{tables/ensemble.tex}
\input{tables/reverie.tex}

\subsection{Pretraining with \airbnb}
\label{sec:xp_pretrain_airbert}
We perform ablation studies on the impact of various methods for creating path-instruction pairs.
We also present ablation studies that highlight the impact of using the shuffling loss during \airbert's pretraining as well as fine-tuning stages.
Throughout this section, our primary focus is on the SR on the unseen validation set and we compare our results against \vlnbert~\cite{majumdar2020vlnbert}, which achieves a SR of 59.26\%.


\p{1. Impact of creating path-instruction pairs.}
Table~\ref{tab:how} presents the performance of multiple ways of using the \airbnb~dataset after ConCaps pretraining as illustrated in Fig.~\ref{fig:model}.
In row 1, we show that directly using \airbnb~IC pairs without any strategies to reduce domain gap improves performance over \vlnbert~by 3.2\%.
Even if we skip ConCaps pretraining, we achieve 60.54\% outperforming 59.26\% of \vlnbert.
It proves that our \airbnb~dataset is more beneficial to VLN than the generic ConCaps dataset.

Naive concatenation (row 2) does only slightly better than using the IC pairs (row 1) as there are still domain shifts with respect to fluency of transitions and lack of visual context.
Rows 3-6 show that each method mitigates domain-shift to some extent.
Instruction rephrasing (row 3) performs better at improving instructions than instruction generation (row 4), possibly since the generator is unable to use the diverse vocabulary of the \airbnb~captions.
Inserting captionless images at random locations (row 6) reduces the domain-shift significantly and achieves the highest individual performance.
Finally, a combination of instruction rephrasing, image merging and captionless insertion provides an overall 3.8\% improvement over concatenation, and a large 7.2\% improvement over \vlnbert.

\p{2. Shuffling loss applied during pretraining.} 
Table~\ref{tab:shuffle} demonstrates that shuffling is an effective strategy to train the model to reason about temporal order, and enforce alignment between PI pairs.
Rows 3-5 show that shuffling is beneficial both during pretraining with \airbnb-PI data, or during fine-tuning with R2R data, and results in 2.3\% and 0.4\% improvements respectively.
In combination with the \emph{Speaker} dataset (paths from seen houses with generated instruction yielding 178K additional PI pairs~\cite{tan2019envdrop}), we see that the shuffling loss provides 3.1\% overall improvement (row 6 vs. 7).
The \airbnb-PI data brings more improvements than the Speaker dataset (row 2 vs. 5).
Putting together the \airbnb-PI data, Speaker dataset and shuffling, we achieve 68.67\% SR on the R2R dataset with a single model.

\p{3. Shuffling loss applied during fine-tuning.}
The final stage of model training on R2R involves fine-tuning to rank multiple candidate paths that form the path selection task.
We compare various approaches to improve this fine-tuning procedure (results in Table~\ref{tab:dataaug-r2r}).
(1) In row 2, we explore the impact of using additional negative paths.
Unsurprisingly, this does not improve performance.
(2) Inspired by~\cite{gupta2020contrastive}, we highlight keywords in the instruction using a part-of-speech tagger~\cite{joshi2018parser}, and include an extra loss term that encourages the model to pay attention to their similarity scores (row 3).
(3) Another alternative suggested by~\cite{gupta2020contrastive} involves masking keywords in the instruction and using VLP models to suggest replacements, resulting in hard negatives (row 4).

Hard negatives and highlighting keywords improve performance by 2.1-2.3\%, but at the cost of extra parsers or VLP models.
In contrast, shuffling visual paths to create two additional negatives results in highest improvement (row 5, +2.7\% on val unseen) and appears to be a strong strategy to enforce temporal order reasoning, that neither requires external parsers nor additional VLP models.

\p{4. Error analysis.}
We study the areas in which \airbert~brings major improvements by analyzing scores for aligned PI pairs and simple corruptions that involve replacing noun phrases (\eg~\emph{bedroom} by \emph{sofa}), swapping noun phrases appearing within the instruction, or switching left and right directions (\eg~\emph{turn left/right} or \emph{leftmost/rightmost chair}).
In particular, for every ground-truth aligned PI pair, we create 10 additional negatives by corrupting the instruction, and measure the accuracy of the model selecting the correct pair.
Table~\ref{tab:analysis} shows that \airbert~with in-domain training and the shuffling loss achieves large improvements ($>$ 8\%) for corruptions involving replacement or swapping of noun phrases.
On the other hand, distinguishing directions continues to be a challenging problem; but here as well we see \airbert~outperform \vlnbert~by 4.5\%.

\subsection{Comparison against state-of-the-art}
\label{sec:xp_sota}
\input{tables/testset.tex}

\p{R2R.}
We first evaluate the discriminative model for the R2R task.
Similar to \vlnbert, we evaluate \airbert~as an ensemble model created by a linear combination (chosen through grid search) of multiple model outputs (see Table~\ref{tab:ensemble}).
First, we see that \airbert~alone (row 2) outperforms \vlnbert~(row 1) by 9.4\% on the unseen environments and a strong ensemble of speaker and follower models~\cite{tan2019envdrop} (row 3) by 0.7\%.
Ensembling \airbert~results in a gain of 1.4\% over the \vlnbert~ensemble (row 4 vs. 5).

We also obtain results on the test set by submitting our best method to the R2R leaderboard%
\footnote{\url{https://eval.ai/web/challenges/challenge-page/97/overview} also shows performance for ensembling \airbert, \vlnbert, speaker and follower at a unseen test set SR of 78\%.}.
As seen from Table~\ref{tab:testset}, our method of ensembling \airbert, speaker, and follower (similar to \vlnbert~with speaker and follower~\cite{devlin2018bert}) achieves the highest success rate at 77\% and is ranked first as of the submission deadline.
Both \vlnbert~and \airbert~use 30 candidate trajectories sampled by beam search with EnvDrop~\cite{tan2019envdrop}, inducing the same path length (PL) for the three methods. As the SPL metric on the leaderboard takes into account the total path length over the 30 trajectories, the SPL is very low and similar across the approaches. 
\airbert~also benefits generative models for the R2R task. The results are presented in the Appendix~\ref{sec:supp-results}.

\p{REVERIE.}
Table~\ref{tab:reverie_results} presents results for the REVERIE dataset.
The last four rows in the table use Recurrent \vlnbert~\cite{hong2021recurrentvln} with different backbones or parameter initialization.
The OSCAR and \vilbert~backbones are pretrained on out-of-domain image-caption pairs.
As compared to OSCAR, we observe slight improvements using the \vilbert~backbone for the REVERIE task.
\vlnbert~shares the same architecture as \vilbert, but is pretrained on the R2R dataset, resulting in performance improvement on the unseen environments.
Our pretrained \airbert~achieves significantly better performance than \vlnbert, with over 2.4\% gain on navigation SR and 1.8\% gain on RGS in unseen environments (val unseen).
Without any special adaptation, we see that \airbert~brings benefits from pretraining on the \airbnb~dataset.
We also achieve the state-of-the-art performance on the REVERIE test set by the time of submission, surpassing previous works by a large margin.

\subsection{Training a navigation agent on few houses}
\label{sec:eval:fsl}
\input{tables/fsl.tex}

We hypothesize that in-domain pretraining, especially one that leverages proposed PI pair generation methods, can achieve superior performance while requiring less training data.
To evaluate this, we propose a novel few shot evaluation paradigm for VLN: models are allowed to fine-tune on samples (PI pairs) from one (or few) environments.
Few-shot learning for VLN is particularly interesting as visual appearance of houses may differ vastly across geographies, and while training data is hard to obtain, pretraining data like \airbnb~may be readily available.


\p{One/few shot tasks.}
We considered two types of setups:
(1) learning from a single environment, which we refer as one-shot learning; and
(2) learning from 6 environments (representing $10\%$ of the total training size).
For both cases, we randomly sample 5 sets of environments, and report average results (standard deviations in the Appendix~\ref{sec:supp-results}).
As the number of paths in an environment may have a large impact on performance, we exclude 17 of 61 environments with less than 80 paths.

\p{Results.}
We adopt \vlnbert, pretrained on ConCaps, as a baseline for few-shot tasks.
Recall that fine-tuning \vlnbert~and \airbert~on R2R relies on candidate paths drawn from an existing model (EnvDrop~\cite{tan2019envdrop}).
However, as this would lead to unfair comparisons (EnvDrop is trained on the full dataset), candidate paths are sampled as the shortest path between two random positions.


Table~\ref{tab:fsl} shows that \airbert~largely outperforms \vlnbert~on the unseen validation set: 27.6\% with 1 house and 22\% with 6 houses.
\airbert~fine-tuned on 6 houses is almost as good as \vlnbert~on the entire training set.
The last two rows of the table shows that using random paths does not lead to a large performance drop for both models and is a testament to the power of pretrained networks.

%% file: tables/how.tex
\begin{table}[t]
\small
\centering
\tabcolsep=0.18cm
\begin{tabular}{l ccccc | cc}
\toprule
& \multirow{2}{*}{Cat} & \multicolumn{2}{c}{Instruction} & \multicolumn{2}{c|}{Path} & \multicolumn{2}{c}{SR on Val} \\
& & Rep & Gen & Merge & Insert & Seen & Unseen \\
\midrule
1 & - & - & - & - & - & 71.21 & 62.45 \\
2 & \checkmark & - & - & - & - & 73.84 & 62.71 \\
3 & - & \checkmark & - & - & - & 72.67 & 63.35 \\
4 & - & - & \checkmark & - & - & 71.19 & 63.11 \\
5 & - & - & - & \checkmark & - & 70.51 & 64.07 \\
6 & - & - & - & - & \checkmark & 74.43 & 66.05 \\
7 & - & \checkmark & - & \checkmark & \checkmark 
                              & 73.57 & \textbf{66.52} \\
\bottomrule
\end{tabular}
\vspace{-2mm}
\caption{Comparison between various \airbnb~PI pair creation strategies for pretraining.
The first row denotes the use of image-caption pairs.
All methods from the second row use masking and shuffling during pretraining.
Cat: naive concatenation;
Rep: instruction rephrasing; 
Gen: instruction generation;
Merge: image merging; and
Insert: captionless image insertion.}
\vspace{-2mm}
\label{tab:how}
\end{table}

%% file: tables/shuffle.tex
\begin{table}[t]
\centering
\small
\tabcolsep=0.09cm
\begin{tabular}{l  cc cc cc | cc}
\toprule
  & 
  \multicolumn{2}{c}{\airbnb} &
  \multicolumn{2}{c}{Speaker} &
  \multicolumn{2}{c|}{R2R} &
  \multicolumn{2}{c}{SR on Val} \\ 
  & Mask & Shuf. & Rank & Shuf. & Rank & Shuf. & Seen & Unseen
  \\ 
  \midrule
1 &  - & - & - & - & \checkmark & - & 70.20 & 59.26  \\
2 &  - & - & \checkmark & \checkmark & \checkmark & \checkmark & 73.12 & 65.50  \\
\midrule
3  &  \checkmark & - & - & - & \checkmark & -  
    & 73.24 & 64.21  \\
4 &  \checkmark & \checkmark &  - & - & \checkmark & -
    & 73.57 & 66.52  \\
5 &  \checkmark & \checkmark &  - & - & \checkmark & \checkmark  
    & 74.69 & 66.90  \\
\midrule
6 & \checkmark & - &  \checkmark & - & \checkmark & -  
    & 70.21 & 65.52 \\
7 & \checkmark & \checkmark &  \checkmark & \checkmark & \checkmark & \checkmark  
    & 73.83 & \textbf{68.67} \\
    \bottomrule
\end{tabular}
\vspace{-2mm}
\caption{
Impact of shuffling during pretraining and fine-tuning.
While additional data helps, we see that using the shuffling loss (abbreviated as Shuf.) consistently improves model performance.
Row 1 corresponds to \vlnbert~\cite{majumdar2020vlnbert}.
}
\vspace{-2mm}
\label{tab:shuffle}
\end{table}

%% file: tables/dataaug-r2r.tex
\begin{table}[t]
\centering
\small
\tabcolsep=0.10cm
\begin{tabular}{ll c | cc}
\toprule
  & Fine-tuning & Additional &
  \multicolumn{2}{c}{SR on Val} \\
   & Strategies & Negatives & Seen & Unseen
  \\   \midrule
  1 & VLN-BERT ~\cite{majumdar2020vlnbert} & 0
    & 70.20 & 59.26 \\
  2 & (1) + Wrong trajectories & 2
    & 70.11 & 59.11 \\
  3 & (1) + Highlight keywords & 0
    & 71.89 & 61.37 \\
  4 & (1) + Hard negatives & 2
    & 71.89 & 61.63 \\
  5 & (1) + Shuffling (Ours) & 2
    & 72.46 & \textbf{61.98} \\
\bottomrule
\end{tabular}
\vspace{-2mm}
\caption{Comparison between different strategies for fine-tuning a \vilbert~model on the R2R task.
\vlnbert~\cite{majumdar2020vlnbert} fine-tunes \vilbert~with a masking and ranking loss.
Each row (described in the text) is an independent data augmentation and can be compared directly against the baseline (row 1).
}
\vspace{-4mm}
\label{tab:dataaug-r2r}
\end{table}

%% file: tables/analysis.tex
\begin{table}[t]
\centering
\small
\tabcolsep=0.08cm
\begin{tabular}{c cc cc cc}
\toprule
\multirow{2}{*}{Model} &
  \multicolumn{2}{c}{Replace-Nouns} &
  \multicolumn{2}{c}{Swap-Nouns} &
  \multicolumn{2}{c}{Directions} \\
  & Seen & Unseen
  & Seen & Unseen
  & Seen & Unseen \\
\midrule
\vlnbert
    & 60.3 & 58.7
    & 53.4 & 52.3
    & 46.2 & 45.3 \\
    
\airbert
    & 68.3 & 66.6
    & 66.6 & 61.1
    & 47.3 & 49.8 \\
\bottomrule
\end{tabular}
\vspace{-2mm}
\caption{Accuracy of models attempting to pick the correct PI pair
from a pool of correct + 10 negatives created by simple corruptions such as replacing or swapping noun phrases and switching directions (left with right).
Random performance is $\frac{1}{11}$ or 9.1\%.}
\vspace{-2mm}
\label{tab:analysis}
\end{table}

%% file: tables/ensemble.tex
\begin{table*}[t]
\small
\tabcolsep=0.10cm
\centering
\begin{tabular}{@{\extracolsep{4pt}}l cccc ccccc ccccc@{}}
\toprule
  & 
  \multirow{2}{*}{\airbert} &
  \vlnbert & Speaker & Follower &
  \multicolumn{5}{c}{Val Seen} &
  \multicolumn{5}{c}{Val Unseen} \\ 
\cline{6-10} \cline{11-15}
  & & \cite{majumdar2020vlnbert} & \cite{tan2019envdrop} & \cite{tan2019envdrop} &
  PL & NE & SPL & OSR & SR & 
  PL & NE & SPL & OSR & SR \\ 
  \midrule
1 & - & \checkmark & - & - &
    10.28 & 3.73 &  0.66 & 76.47 & 70.20 &
     9.60 & 4.10 &  0.55 & 69.22 & 59.26 \\
2 & \checkmark & - & - & - &
    10.59 & 3.21 & 0.69 & 80.71 & \textbf{73.85} & 
    10.03 & 3.24 & 0.63 & 78.45 & \textbf{68.67} \\
3 & - & - & \checkmark & \checkmark &
    10.69 & 2.72 &  0.70 & 82.94 & 74.22 &
    10.10 & 3.32 &  0.63 & 76.63 & 67.90 \\
\midrule
4 & - & \checkmark & \checkmark  & \checkmark &
    10.61 & 2.35 &  0.78 & 86.57 & \textbf{81.86} &
    10.00 & 2.76 &  0.68 & 81.91 & 73.61 \\
5 & \checkmark & - & \checkmark & \checkmark &
    10.63 & 2.13 & 0.77 & 87.17 & \textbf{81.40} &
     9.99 & 2.69 & 0.70 & 82.89 & \textbf{75.01} \\
\bottomrule
\end{tabular}
\vspace{-2mm}
\caption{Performance of single models and the impact of ensembling \vlnbert~or \airbert~with the speaker and follower.}
\vspace{-2mm}
\label{tab:ensemble}
\end{table*}

%% file: tables/reverie.tex
\begin{table*}
\footnotesize
\centering
\tabcolsep=0.08cm
\begin{tabular}{l|cccccc|cccccc|cccccc} \toprule
\multirow{3}{*}{Methods} & \multicolumn{6}{c|}{Validation Seen} & \multicolumn{6}{c|}{Validation Unseen} & \multicolumn{6}{c}{Test Unseen} \\
\multicolumn{1}{c|}{} & \multicolumn{4}{c}{Navigation} & \multirow{2}{*}{RGS} & \multirow{2}{*}{RGSPL} & \multicolumn{4}{c}{Navigation} & \multirow{2}{*}{RGS} & \multirow{2}{*}{RGSPL} & \multicolumn{4}{c}{Navigation} & \multirow{2}{*}{RGS} & \multirow{2}{*}{RGSPL} \\ 
& SR & OSR & SPL & TL &  &  & SR & OSR & SPL & TL &  &  & SR & OSR & SPL & TL &  &  \\ \midrule
Seq2Seq-SF \cite{anderson2018r2r} & 29.59 & 35.70 & 24.01 & 12.88 & 18.97 & 14.96 & 4.20 & 8.07 & 2.84 & 11.07 & 2.16 & 1.63 & 3.99 & 6.88 & 3.09 & 10.89 & 2.00 & 1.58 \\
RCM \cite{wang2019reinforced} & 23.33 & 29.44 & 21.82 & 10.70 & 16.23 & 15.36 & 9.29 & 14.23 & 6.97 & 11.98 & 4.89 & 3.89 & 7.84 & 11.68 & 6.67 & 10.60 & 3.67 & 3.14 \\
SMNA \cite{ma2019self} & 41.25 & 43.29 & 39.61 & 7.54 & 30.07 & 28.98 & 8.15 & 11.28 & 6.44 & 9.07 & 4.54 & 3.61 & 5.80 & 8.39 & 4.53 & 9.23 & 3.10 & 2.39 \\
FAST-MATTN \cite{qi2020reverie} & \textbf{50.53} & \textbf{55.17} & \textbf{45.50} & \textbf{16.35} & 31.97 & 29.66 & 14.40 & 28.20 & 7.19 & 45.28 & 7.84 & 4.67 & 19.88 & 30.63 & 11.61 & 39.05 & 11.28 & 6.08 \\
Rec (OSCAR) \cite{hong2021recurrentvln} & 39.85 & 41.32 & 35.86 & 12.85 & 24.46 & 22.28 & 25.53 & 27.66 & 21.06 & 14.35 & 14.20 & 12.00 & 24.62 & 26.67 & 19.48 & 14.88 & 12.65 & 10.00 \\ \midrule
Rec (\vilbert) & 43.64 & 45.61 & 37.86 & 15.75 & 31.69 & 27.58 & 24.57 & 29.91 & 19.81 & 17.83 & 15.14 & 12.15 & 22.17 & 25.51 & 17.28 & 18.22 & 12.87 & 10.00 \\
Rec (\vlnbert) & 41.11 & 42.87 & 35.55 & 15.62 & 28.39 & 24.99 & 25.53 & 29.42 & 20.51 & 16.94 & 16.42 & 13.29 & 23.57 & 26.83 & 18.73 & 17.63 & 14.24 & 11.63 \\
Rec (\airbert) & 47.01 & 48.98 & 42.34 & 15.16 & \textbf{32.75} & \textbf{30.01} & \textbf{27.89} & \textbf{34.51} & \textbf{21.88} & \textbf{18.71} & \textbf{18.23} & \textbf{14.18} & \textbf{30.28} & \textbf{34.20} & \textbf{23.61} & \textbf{17.91} & \textbf{16.83} & \textbf{13.28} \\ \bottomrule
\end{tabular}
\vspace{-2mm}
\caption{Navigation and object localization performance on the REVERIE dataset, including results on the unseen test set (leaderboard).}
\vspace{-4mm}
\label{tab:reverie_results}
\end{table*}

%% file: tables/testset.tex
\begin{table}[t]
\centering
\small
\tabcolsep=0.13cm
\begin{tabular}{l ccccc}
\toprule
\multirow{2}{*}{Model} & \multicolumn{5}{c}{Test Unseen} \\ \cline{2-6}
 & PL & NE & SPL & OSR & SR \\
\midrule
Speaker-Follower~\cite{fried2018speaker} & 1,257 & 4.87 & 0.01 & 96 & \h 53 \\
PreSS~\cite{li2019press} & 10.5 & 24.5 & 0.63 & 57 & \h 53  \\
PREVALENT~\cite{hao2020prevalent} & 10.21 & 4.52 & 0.56 & 64 & \h 59 \\
Self-Monitoring~\cite{ma2019self} & 373 & 4.48 & 0.02 & 97 & \h 61 \\
Reinforced CM~\cite{wang2019reinforced} & 358 & 4.03 & 0.02 & 96 & \h 63 \\
EnvDrop~\cite{anderson2018r2r} & 687 & 3.26 & 0.01 & 99 & \h 69 \\
AuxRN~\cite{zhu2020auxrn} & 41 & 3.24 & 0.21 & 81 & \h 71 \\
\vlnbert~\cite{majumdar2020vlnbert} & 687 & 3.09 & 0.01 & 99 &  \h73 \\
\midrule
\airbert~(ours) & 687 & 2.69 & 0.01 & 99 & \h 77 \\
\bottomrule
\end{tabular}
\vspace{-2mm}
\caption{Navigation performance on the R2R unseen test set as indicated on the benchmark leaderboard.}
\vspace{-2mm}
\label{tab:testset}
\end{table}


%% file: tables/fsl.tex
\begin{table}[t]
\centering
\small
\tabcolsep=0.14cm
\begin{tabular}{ccc cc cc}
\toprule
\multirow{2}{*}{\# Env.} &
\multirow{2}{*}{Traj.} & 
\multicolumn{2}{c}{\vlnbert~SR} &
\multicolumn{2}{c}{\airbert~SR}\\ 
& & Seen & Unseen & Seen & Unseen \\ 
\midrule
1  & Rand 
    & 45.71 & 22.43
    & 47.88 & 50.00 \\
\midrule
6  & Rand
    & 52.75 & 35.99
    & 54.48 & 57.97 \\
\midrule
61 & Rand
    & 67.68 & 57.15
    & 64.24 & 65.60 \\
61 & \cite{tan2019envdrop}
    & 70.20 & 59.26
    & 73.83 & 68.48 \\
\bottomrule
\end{tabular}
\vspace{-2mm}
\caption{Performance on R2R few-shot evaluation.
During training, only a subset of the Matterport~\cite{Matterport3D} environments are accessible.
Standard deviation is reported in the supplementary material.}
\vspace{-4mm}
\label{tab:fsl}
\end{table}


%% file: sections/7.conclusion.tex
\section{Conclusion}
\label{sec:conclusion}
We introduced \airbnb, a large-scale, in-domain, image-text dataset from houses listed on online rental marketplaces and showed how domain gaps between \airbnb~image-caption pairs and VLN tasks can be mitigated through the creation of path-instruction pairs.
We also proposed shuffling, as a means to improve an agent's reasoning about temporal order.
Our pretrained model \airbert, achieved state-of-the-art on R2R through the discriminative path-selection setting, and REVERIE through a generative setting.
We also demonstrated large performance improvements when applying our model to a challenging one/few-shot VLN setup, highlighting the impact of good pretraining in VLN tasks.






%% file: sections/supp.tex
In this supplementary material, we present additional details, statistics and examples for the \airbnb~dataset; we discuss implementation details for the models used in our work; and present qualitative results as well as the detailed results for the new few-shot learning paradigm.

\section{\airbnb~dataset}
\label{sec:supp-bnb}

This section presents additional details for our \emph{Bed-and-Breakfast} (BnB) dataset.
We start by a short discussion of image-caption pairs (\airbnb~IC) collected from an online rental marketplaces and their statistics.
Subsequently, we present how a combinatorially large number of path-instruction pairs (\airbnb~PI) can be created automatically.
We end this section with multiple examples of \airbnb~PI pairs generated via the concatenation and domain-shift reduction (\eg~rephrasing, captionless insertion) strategies.

\subsection{Filtering image-caption pairs: Outdoor images}
\begin{figure*}[t]
\centering
\begin{subfigure}[b]{0.30\textwidth}
\centering 
\includegraphics[width=\textwidth]{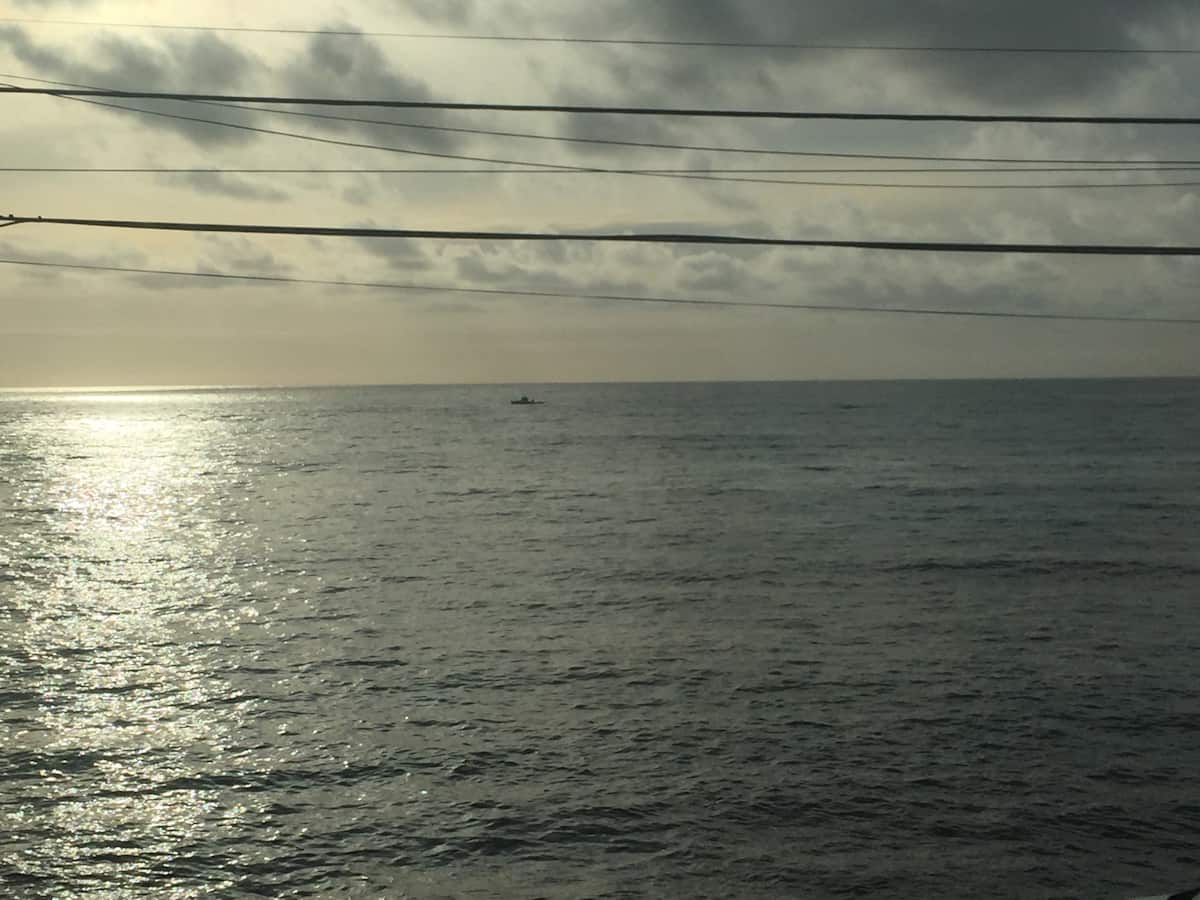}
\caption{Gorgeous ocean views from {\color{red}bedroom}.}
\label{fig:view}
\end{subfigure}\quad
\begin{subfigure}[b]{0.339\textwidth}
\centering 
\includegraphics[width=\textwidth]{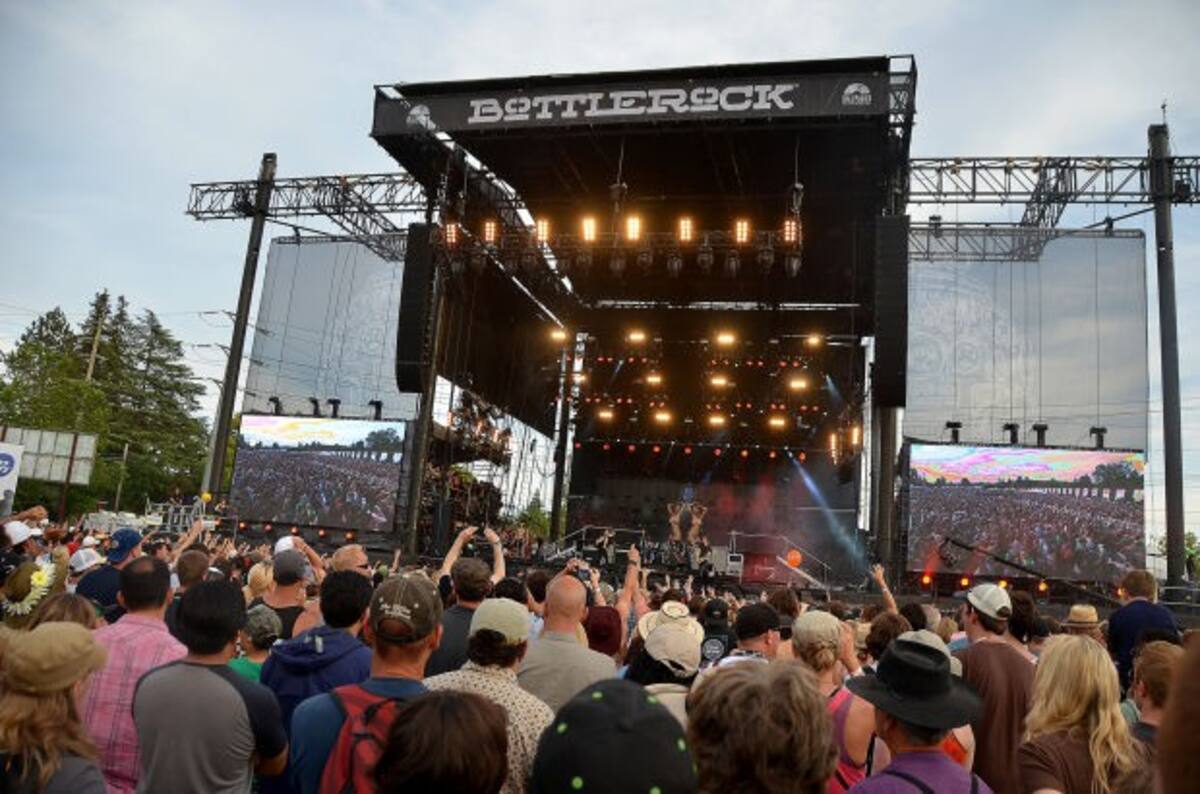}
\caption{Main stage}
\label{fig:festival}
\end{subfigure}\quad
\begin{subfigure}[b]{0.30\textwidth}
\centering 
\includegraphics[width=\textwidth]{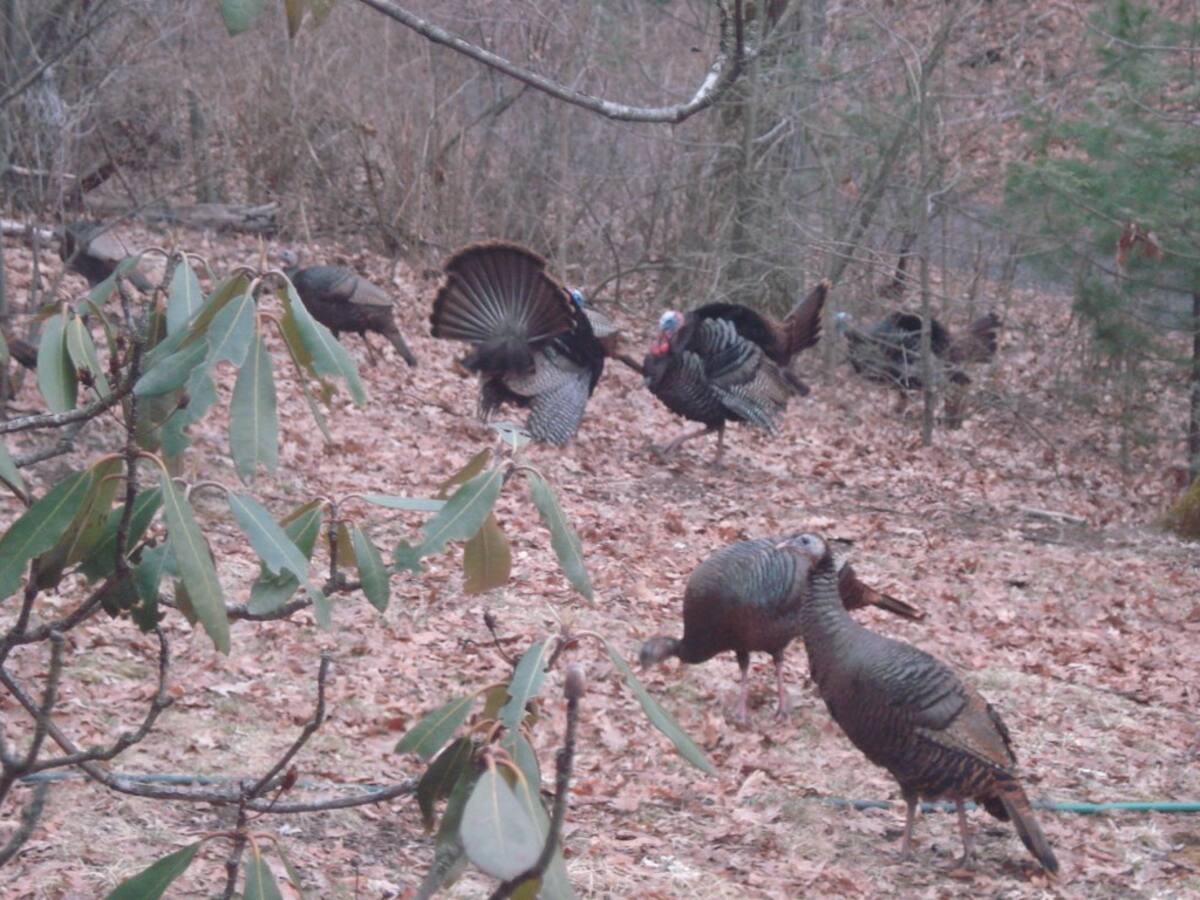}
\caption{Spring excitement}
\label{fig:confusing}
\end{subfigure}
\caption{Examples of outdoor images with their corresponding captions.}
\label{fig:outdoor}
\end{figure*}

Images of outdoor scenes are almost never seen in the environments used in downstream VLN tasks.
In fact, not only are the images out-of-domain (such images are rarely seen in the VLN environments), their captions are often irrelevant to a VLN task.
In order to alleviate the impact of such noisy images and captions, we discard outdoor images from the pretraining process.
Figure~\ref{fig:outdoor} illustrates several examples of misleading outdoor image-caption pairs.
Captions as written by the host are presented in the label below the image.
The caption for the image in Figure~\ref{fig:view} refers to a ``\emph{bedroom}'', however, the image does not show a bedroom.
Similarly, the image-caption pair in the Figure~\ref{fig:festival} talks about activities or festivals that take place in the neighborhood of the listing, however, they are not relevant for solving indoor navigation tasks.
Finally, Figure~\ref{fig:confusing} shows an outdoor scene with several birds along with a noisy caption that is not directly related to the image content, but the emotion that the image may evoke.

\subsection{Dataset details and Statistics}

\begin{figure*}[t]
    \centering
    \begin{subfigure}[b]{0.47\textwidth}
        \centering
        \includegraphics[width=\textwidth]{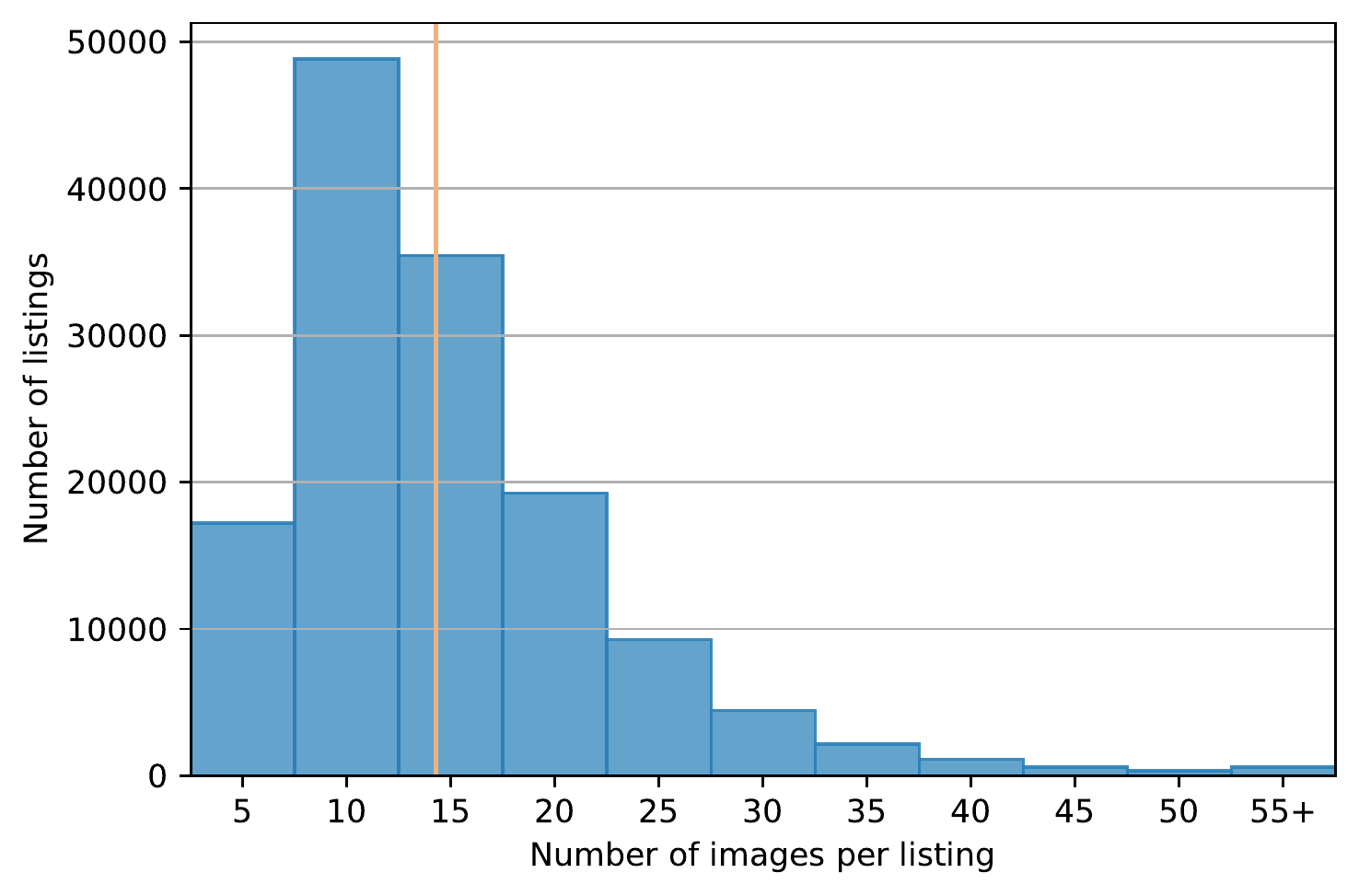}
        \caption{Distribution of the number of images per listing.}
        \label{fig:imgs_per_listing}
    \end{subfigure}
    \hfill
    \begin{subfigure}[b]{0.47\textwidth}
        \centering
        \includegraphics[width=\textwidth]{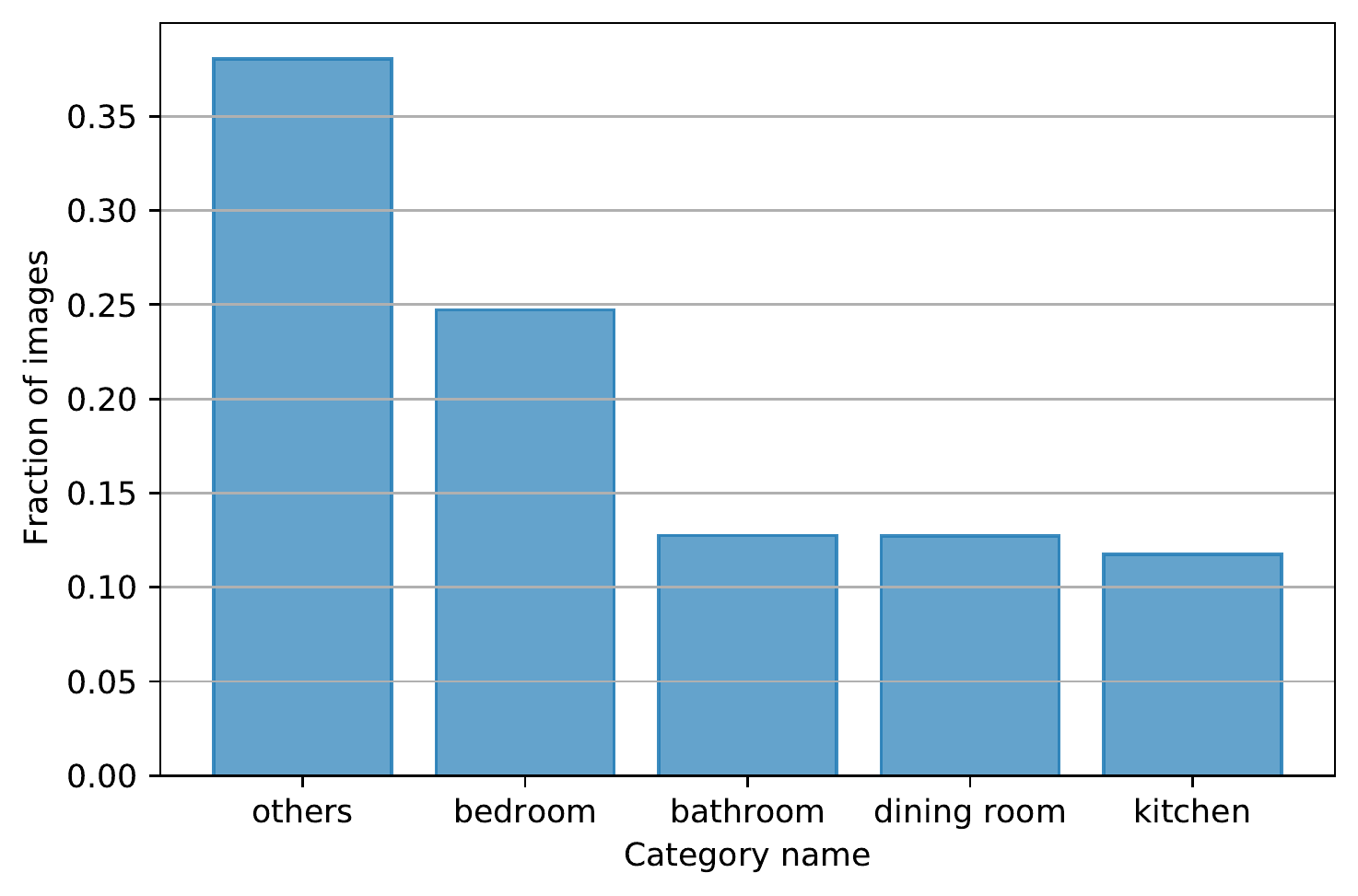}
        \caption{Distribution of predicted scene categories on \airbnb~images.}
        \label{fig:img_categories}
    \end{subfigure}
    \hfill
    \begin{subfigure}[b]{0.47\textwidth}
        \centering
        \includegraphics[width=\textwidth]{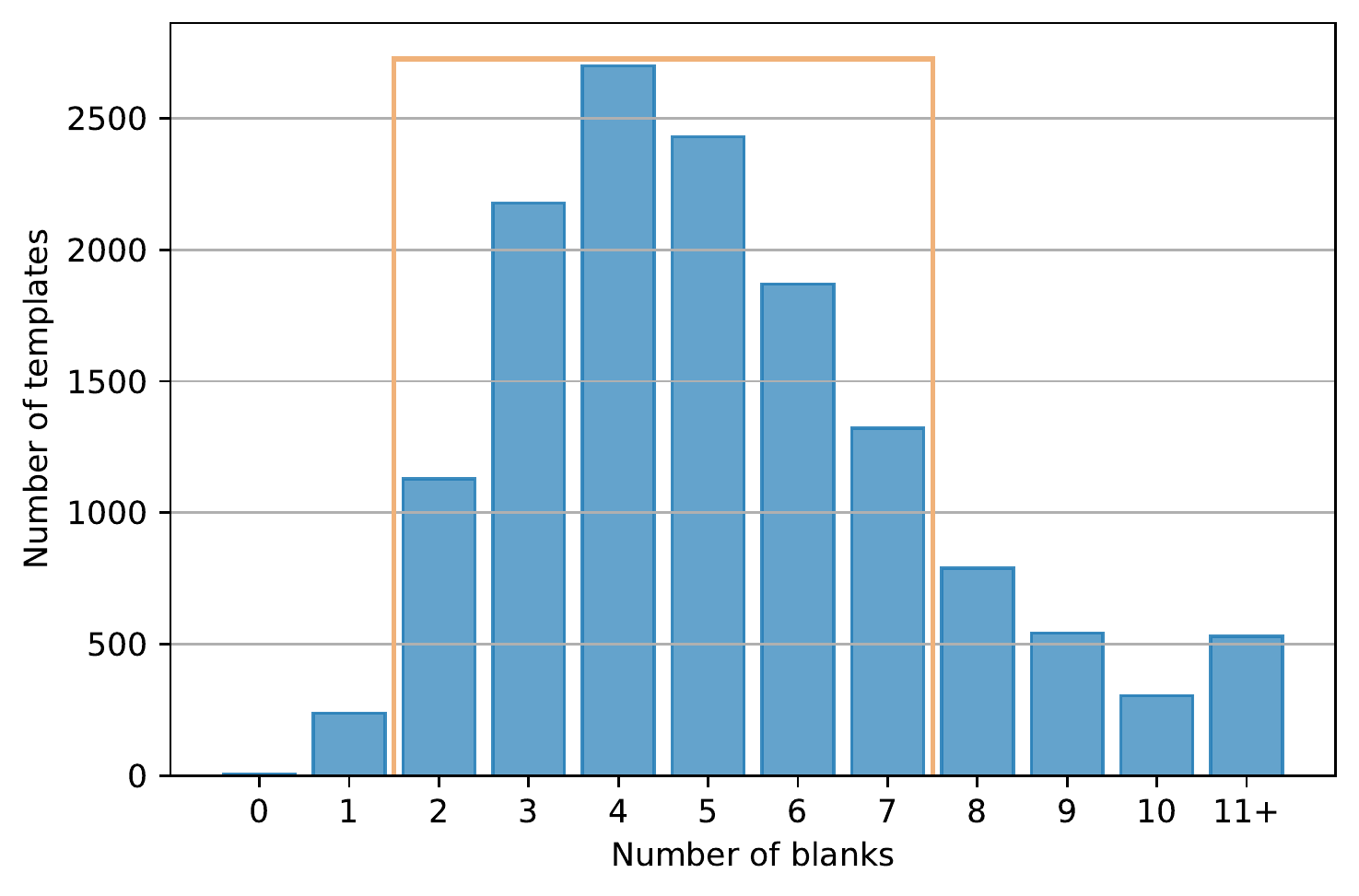}
        \caption{Fill-in-the-blanks templates built using the R2R training set.}
        \label{fig:fillin-templates}
    \end{subfigure}
    \hfill
    \begin{subfigure}[b]{0.47\textwidth}
         \centering
         \includegraphics[width=\textwidth]{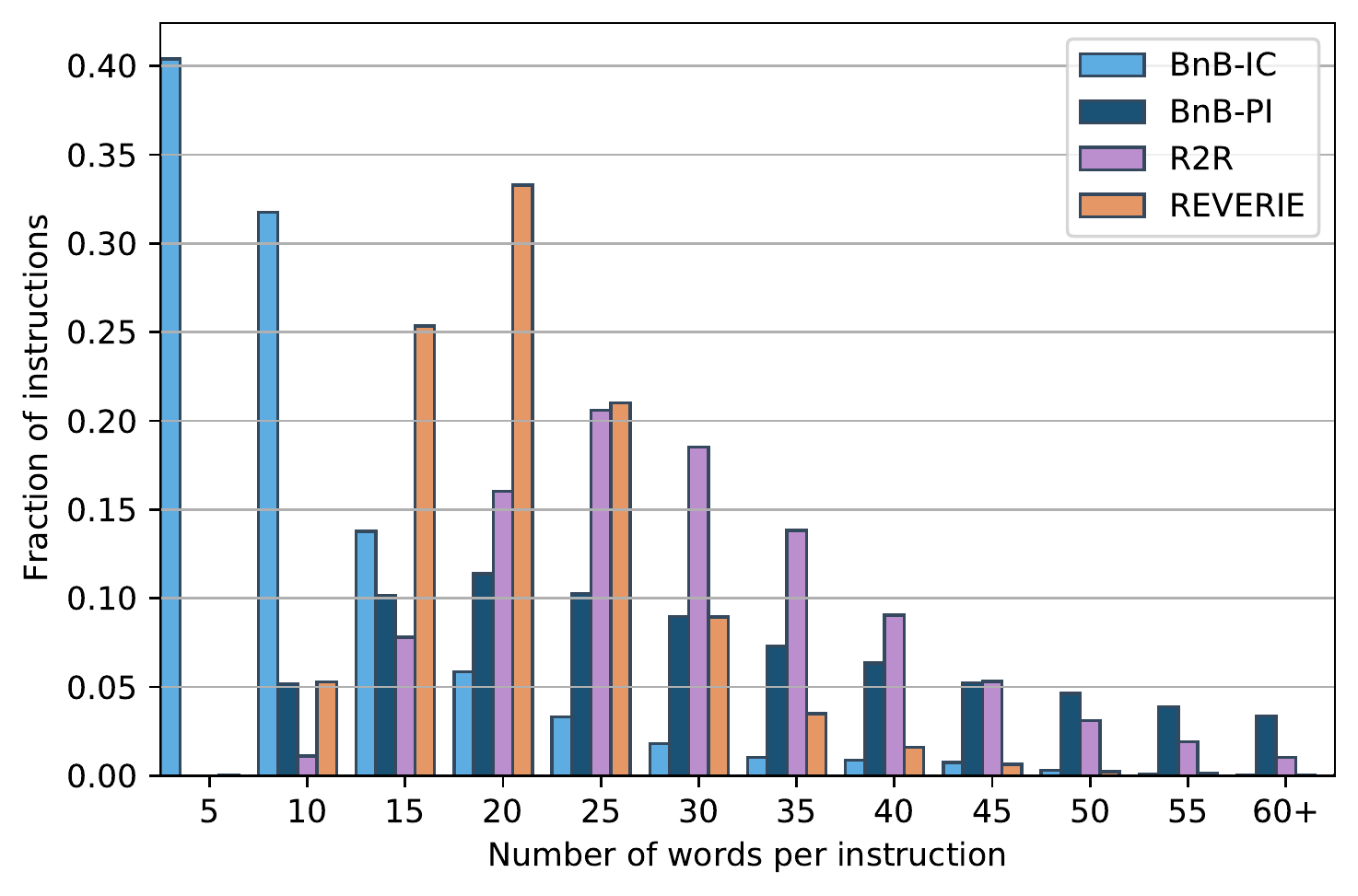}
         \caption{Distribution of the instruction lengths.}
         \label{fig:words_in_caption}
     \end{subfigure}
    \caption{Statistics of \airbnb~Dataset.}
    \label{fig:statistics}
\end{figure*}

\p{\airbnb~image-caption pairs.}
We collect \airbnb~IC pairs from 150K listings on \emph{Airbnb} resulting in 713K image-caption pairs and 676K images without captions.
In Figure~\ref{fig:statistics}, we present some key statistics about this data.
Figure~\ref{fig:imgs_per_listing} presents a histogram of the number of images found in each listing.
While most listings have less than 20 images, this is still a sufficiently large and diverse in-domain distribution.
In Figure~\ref{fig:img_categories}, we summarize the rooms depicted in the images through predicted category labels obtained using a CNN trained on the \emph{Places365} dataset~\cite{zhou2017places}.
These category labels are used as part of our proposed extensions such as \emph{image merging}.

\p{Creating \airbnb~path-instruction pretraining samples.}
We create the \airbnb~PI pairs on-the-fly during training to mimic the agent's visual trajectory and a corresponding instruction through an environment.
Each sample in a batch is created by randomly sampling a listing without replacement during an epoch (one epoch consists of one PI pair from each listing).
Then, the number of IC pairs $K$ that form the PI pair are chosen (as an integer) from a uniform distribution, $K \sim U[4, 7]$.
We sample $N \sim U[2, K]$ IC pairs that have a non-empty caption and the remainder $K-N$ images are chosen from the set of captionless images.
Any image in the path may include additional visual context (from the same room) via the \emph{image merging} strategy.
Similarly, the \emph{instruction rephrasing} strategy may be employed by using existing R2R instruction templates and filling them with noun phrases extracted from the image captions.

The above procedure results in creating one correctly aligned (positive) PI pair, ($X^+$ in the main paper).
To employ the shuffling loss for each sample, we create 9 additional negatives ($X^-_n$ in the paper) by shuffling either the sequence of images or captions, ensuring that the post-shuffling order does not align with the positive pair.

\p{Statistics for \airbnb~PI pairs.}
Due to the large number of possible combinations, we can (theoretically) create 200 billion path-instruction pairs, using the simple concatenation strategy.
This number grows to over 300 quadrillion when considering additional visual context augmentations and fluent instructions.

For \emph{instruction rephrasing}, we create 11,626 fill-in-the-blank templates from the R2R training set.
Figure~\ref{fig:fillin-templates} shows the distribution of the number of blanks in the templates -- most instruction templates have 2-7 blanks into which we insert noun phrases from the \airbnb~captions.

While we are unable to generate the entire \airbnb~PI dataset for computing statistics, we generate 50K PI pairs as a representative sample.
Figure~\ref{fig:words_in_caption} presents the distribution of instruction lengths (number of words) for different datasets.
We see that the captions in \airbnb~IC pairs are much shorter than typical instructions in R2R and REVERIE, while our automatically created instructions in \airbnb~PI pairs exhibit a high level of similarity in terms of their length.



\begin{figure*}[t]
     \centering
     \begin{subfigure}[b]{0.9\textwidth}
         \centering
         \includegraphics[width=\textwidth]{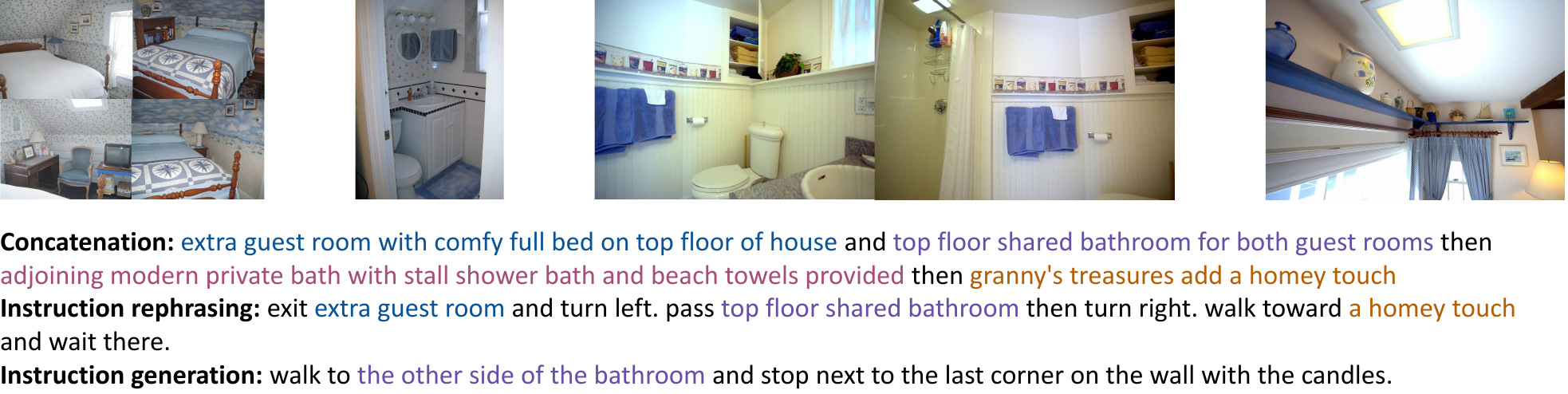}
         \caption{Example 1}
         \label{fig:pi-pair-1}
     \end{subfigure}
     \hfill
     \begin{subfigure}[b]{0.9\textwidth}
         \centering
         \includegraphics[width=\textwidth]{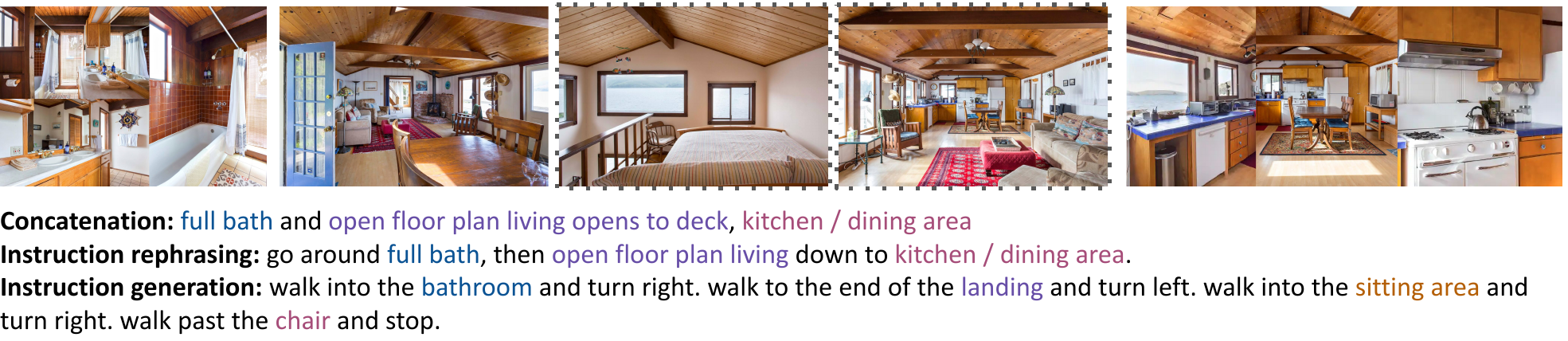}
         \caption{Example 2}
         \label{fig:pi-pair-2}
     \end{subfigure}
     \hfill
     \begin{subfigure}[b]{0.9\textwidth}
         \centering
         \includegraphics[width=\textwidth]{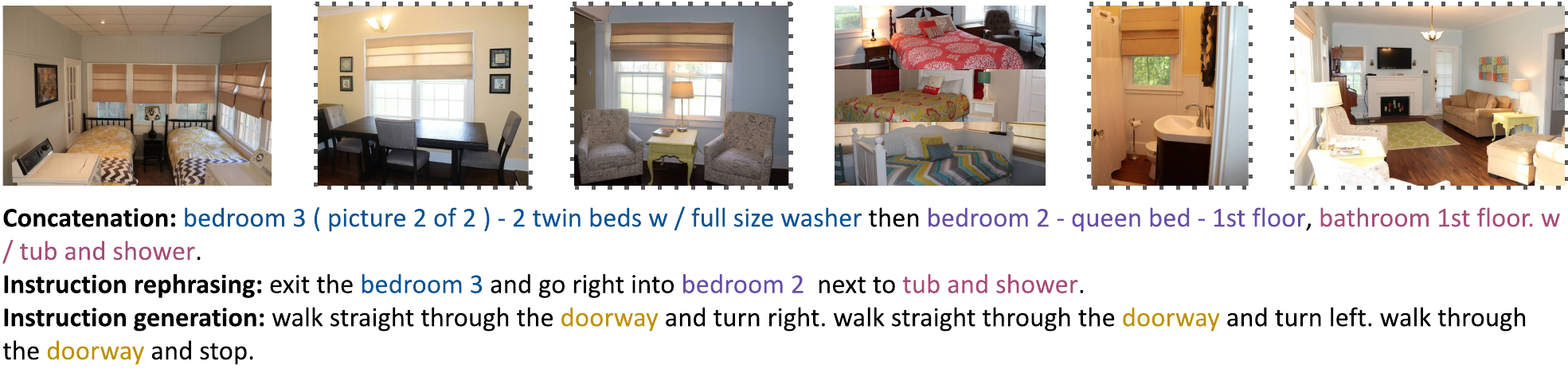}
         \caption{Example 3}
         \label{fig:pi-pair-3}
     \end{subfigure}
    \caption{Examples of path-instruction pairs created by different strategies. The images with dotted borders are images chosen from the \emph{captionless image insertion} strategy, and the clustered images are from the \emph{image merging} strategy.}
    \label{fig:pi-pairs}
\end{figure*}

\subsection{Examples of \airbnb~PI Pairs}
Figure~\ref{fig:pi-pairs} presents generated \airbnb~PI pairs using various strategies proposed in our work, including naive concatenation, instruction rephrasing, instruction generation, image merging and captionless image insertion.

Among the methods to create an instruction, simple concatenation lacks action verbs between sentences for fluent transition leading to a domain shift from real instructions.
Instruction rephrasing selects noun phrases from \airbnb~image descriptions and inserts them into real instruction templates, providing a natural feel to the created instruction.
Finally, while the learning approach of instruction generation (recall, this is learned on downstream VLN dataset) produces fluent sentences, it is unable to leverage the diverse captions of \airbnb~images due to the limited vocabulary stemming from the downstream VLN dataset.
For example, the generated instruction in Figure~\ref{fig:pi-pair-3} does not contain noun phrases related to images in the path.
Better caption generation models such as Pointer network~\cite{vinyals2015pointer}
may help avoid such problems, however are left for future work.

Among augmentations for path generation, we can see that \emph{image merging} helps to expand relevant visual context from single images to semi-panoramic views, see the bedroom in Figure~\ref{fig:pi-pair-1} or the kitchen in Figure~\ref{fig:pi-pair-2}.
\emph{Captionless image insertion} also improves the path diversity by mimicking unmentioned viewpoints in the instruction (indicated by images with a dotted border).

\begin{figure*}[t]
     \centering
     \begin{subfigure}{0.9\textwidth}
         \centering
         \includegraphics[width=\textwidth]{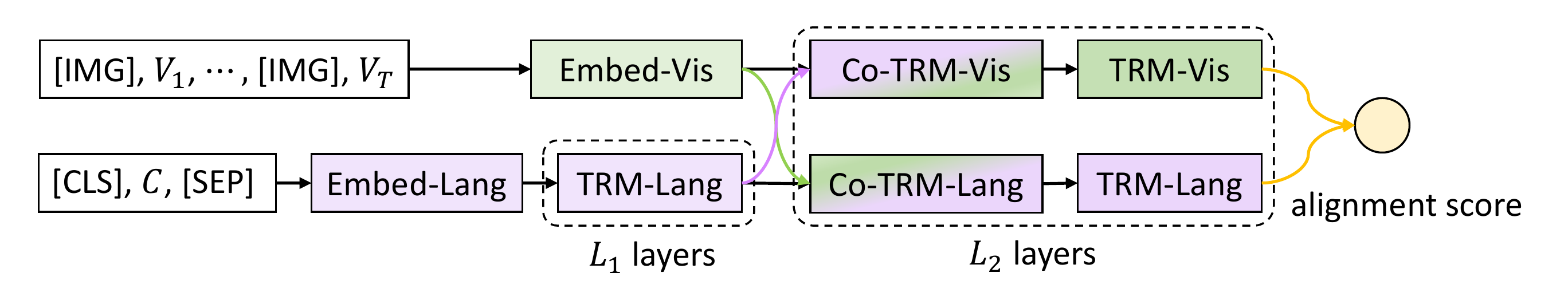}
         \caption{Adapting \airbert~to a discriminative setting to predict path-instruction alignment score, similar to~\cite{majumdar2020vlnbert}.}
         \label{fig:vilbert_disc_model}
     \end{subfigure}
     \hfill
     \begin{subfigure}{0.88\textwidth}
         \centering
         \includegraphics[width=\textwidth]{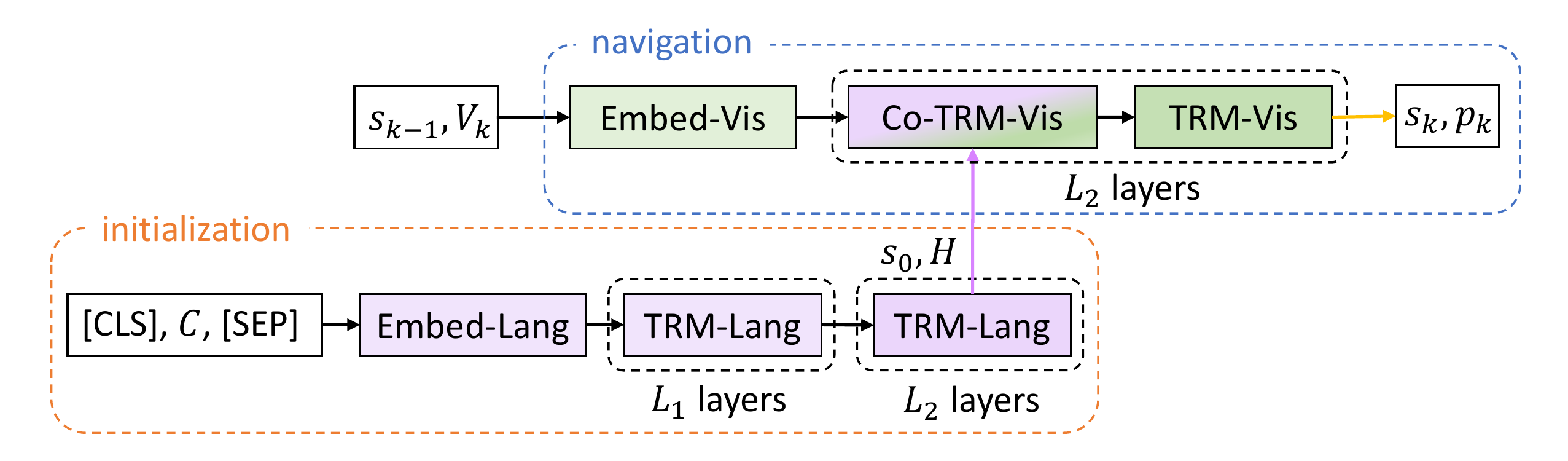}
         \caption{Adapting \airbert~to a generative setting based on the Recurrent VLN-BERT~\cite{hong2021recurrentvln}.}
         \label{fig:vilbert_gen_model}
     \end{subfigure}
    \caption{The adapted \airbert~model in both discriminative and generative settings for downstream VLN tasks.}
    \label{fig:models}
\end{figure*}

\section{Implementation details}
\label{sec:supp-details}

We present the implementation details for learning \airbert~via pretraining using \airbnb, and subsequent fine-tuning in both discriminative or generative settings.

\subsection{\airbert~Pretraining}
\airbert's architecture is the same as \vlnbert~(see Figure~\ref{fig:vilbert_disc_model} where the number of layers $L_1 = L_2 = 6$).
The feature vector $v^k_i$  (corresponding to $i$th image region of the $k$th image) is composed of three terms: the first term is the visual feature extracted by the Bottom-Up Top-Down attention model~\cite{anderson2017butd};
the second term encodes the location of the region in the image as $\text{MLP}(l^k_i)$, where $l^k_i$ is the 5-dim location vector of the given image region defined as the top corner $(x, y)$, the width, height and area;
and the last term $\text{Emb}(k)$ encodes the position, where $\text{Emb}$ is an embedding layer for the image order.

We use 8 V100 SXM2 GPUs (32 GB each) for pretraining~\airbert.
The model is trained for 15 epochs with a batch size of 64 and learning rate of $4\times10^{-5}$.
Each epoch consists of one randomly sampled PI pair from 95\% of the listings, while the remaining 5\% are used for validation and preventing overfitting.

\subsection{Fine-tuning in Discriminative Setting}
In the discriminative setting, R2R navigation is formulated as a path selection problem given the instruction.
The pretrained \airbert~model can be directly fine-tuned without any modifications to the architecture to predict the path-instruction alignment (or compatibility) score as shown in Figure~\ref{fig:vilbert_disc_model}.

We follow the same fine-tuning setup as \vlnbert~\cite{majumdar2020vlnbert} to allow for a fair comparison.
We use the Adam optimizer with a learning rate of $4\times10^{-5}$.
The optimizer is controlled by a learning rate scheduler with a linear warmup and cooldown. 
We fine-tune \airbert~for 30 epochs with a batch size of 64.
Samples from the R2R training set are used for fine-tuning and the model checkpoint with the highest success rate on the unseen validation set (val unseen) is selected for the test set and leaderboard submission.

\subsection{Fine-tuning in Generative Setting}
In the generative setting, an agent is required to predict navigable actions step by step.
We adopt the state-of-the-art generative model Recurrent VLN-BERT~\cite{hong2021recurrentvln} for R2R and REVERIE tasks.
The model uses a pretrained multimodal transformer as a backbone and adds recurrence to a state token to keep track of history for sequential action prediction.
Although the original Recurrent VLN-BERT model only implements an LXMERT-like~\cite{tan2019lxmert} architecture PREVALENT~\cite{hao2020prevalent}, and one-stream BERT-like architecture OSCAR~\cite{li2020oscar}, it is easy to plug our two-stream \vilbert~architecture as the backbone.

The adapted model is shown in Figure~\ref{fig:vilbert_gen_model}.
For initialization, the language stream is used to encode the instruction $C$ into an instruction representation $H$.
As no visual inputs are used during the initialization, the co-attention modules in the original language stream of \vilbert~are removed, and the output feature of the \cls~token is used as the agent's initial state $s_0$.
For navigation at each step $k$, the visual stream takes the previous state $s_{k-1}$, visual observations $V_k$ at step $k$ and the encoded language features $H$ to generate a new state $s_k$ and action decision $p_k$.

When fine-tuning on the R2R dataset, we use scene features with a ResNet-152 pretrained on Places365~\cite{zhou2017places} and augment the training data with generated path-instruction pairs from~\cite{hao2020prevalent}.
We train the model via imitation learning and A2C reinforcement learning for 300,000 iterations with a batch size of 16 and learning rate of $10^{-5}$.
When fine-tuning on the REVERIE dataset, object features encoded by a Bottom-Up Top-Down attention model~\cite{anderson2017butd} are used along with the scene features.
The model is trained for 200,000 iterations with a batch size of 8.
All the experimental setups for fine-tuning are the same as~\cite{hong2021recurrentvln} for a fair comparison.

\section{Results}
\label{sec:supp-results}

In this section, we present additional results on adapting \airbert~to a generative setting and applying it to the R2R task.
Through several qualitative examples, we obtain a better understanding for \airbert's performance improvements, and finally present detailed results on the new few-shot learning paradigm in VLN.

\subsection{Results on R2R with Generative Models}
\input{tables/r2r_generative}

Table~\ref{tab:r2r_generative_results} shows the performance of different generative models on the R2R dataset.
The OSCAR and \vilbert~backbones for Recurrent VLN-BERT~\cite{hong2021recurrentvln} (Rec) are all pretrained on large-scale out-of-domain image-caption pairs with object features and similar self-supervised tasks.
On the other hand, the PREVALENT~\cite{hao2020prevalent} backbone is pretrained on in-domain R2R dataset with scene features and fine-tuned with an additional action prediction task.
We suspect that this is the reason for PREVALENT's higher performance as compared to using OSCAR or \vlnbert~as backbones.
Note that our \airbert~backbone is not fine-tuned further on downstream tasks after pretraining.

Replacing OSCAR's single BERT-like architecture with the \vilbert~architecture slightly improves the performance (similar to our results on the REVERIE dataset presented in the main paper).
The \vlnbert~model further fine-tunes \vilbert~on the R2R dataset (with the masking loss).
This is beneficial to the navigation performance on the unseen environments validation set\footnote{The performance of \vlnbert~on the seen validation set is lower because the model checkpoint is selected to maximize performance on validation unseen set which happens to be at an earlier iteration.}.
Our \airbert~initialization achieves substantial performance improvement as compared to the OSCAR and \vlnbert~backbones on unseen environments, while achieving comparable performance with the PREVALENT initialization.

\subsection{Qualitative results}
\newcommand{\gfinal}{{\textcolor{SeaGreen}\final}}
\newcommand{\pfinal}{{\color{Orchid}\final}}
\newcommand{\bfinal}{{\color{cyan}\final}}

We visualize the predicted paths from \vlnbert~and \airbert~models.
In the following figures, {\color{yellow}\start{}} is the starting viewpoint of the agent, \gfinal{} denotes viewpoints in the ground-truth path, \pfinal{} for \vlnbert~and \bfinal{} for \airbert. Arrows indicate the navigation direction.

\begin{figure*}[t]
    \centering
    \begin{subfigure}[b]{0.47\textwidth}
        \centering
        \includegraphics[width=\textwidth]{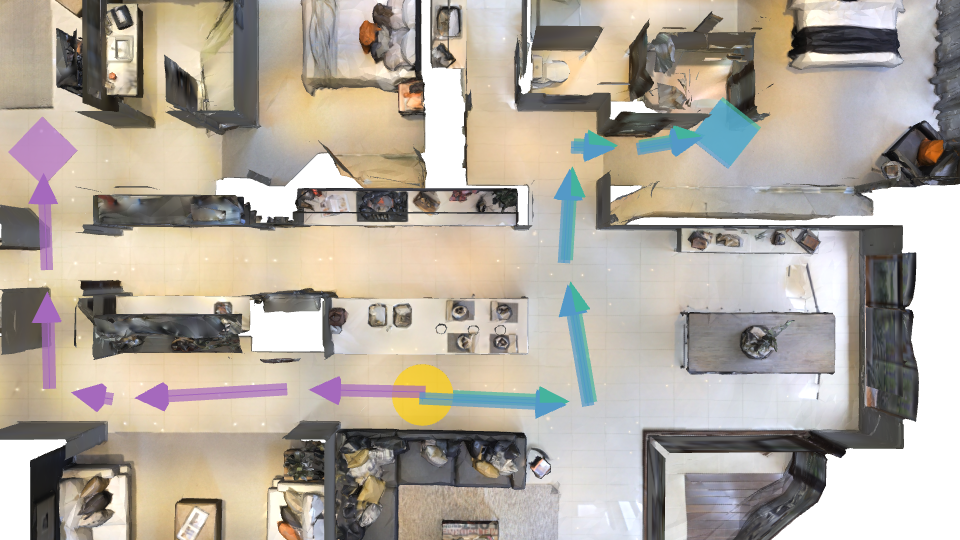} 
        \caption{\textbf{R2R \success{}:} Walk over the kitchen counter, turn left,  walk ahead till wall, turn right, walk to the closet room, wait at front.}%
        \label{fig:r2r-closet}%
    \end{subfigure}
    \hfill
    \begin{subfigure}[b]{0.47\textwidth}
         \centering
          \includegraphics[width=\textwidth]{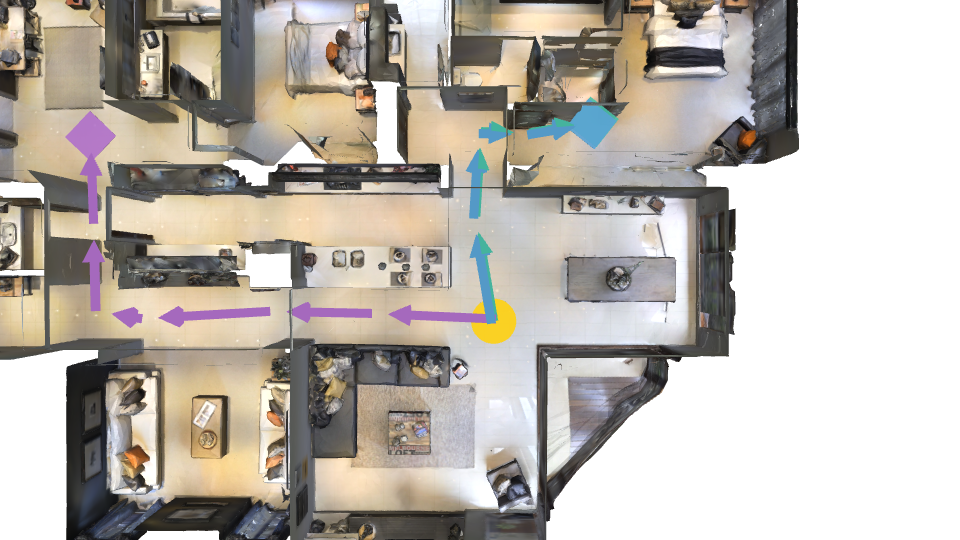} 
          \caption{\textbf{REVERIE \success{}:} Walk past the kitchen and enter the hallway. Turn right at the artwork and wait by the closet.}%
          \label{fig:reverie-closet}%
     \end{subfigure}
    \hfill

    \begin{subfigure}[b]{0.47\textwidth}
        \centering
           \includegraphics[width=\textwidth]{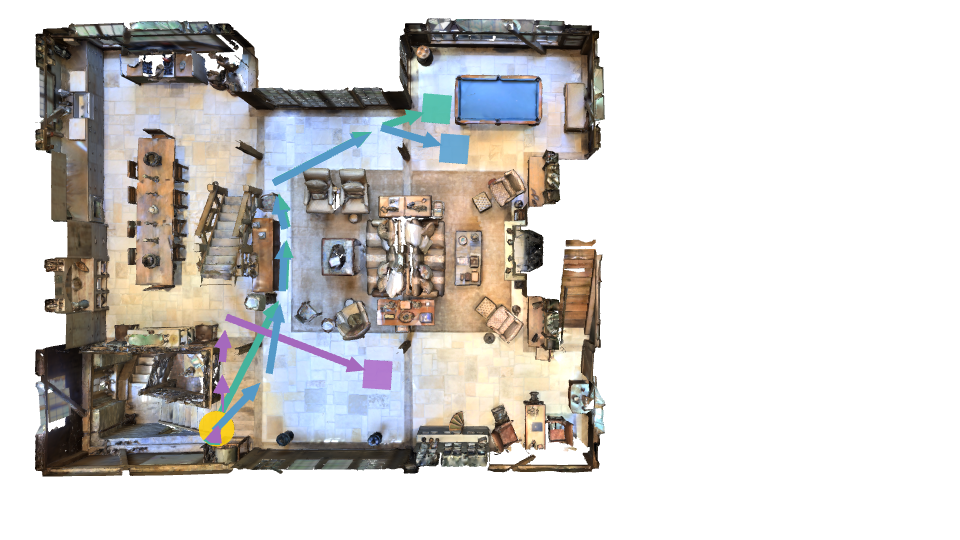}
          \hspace*{-3cm}\caption{\textbf{R2R \success{}:} Walk forward to the sitting area to the right of the stairs. Walk to the wall of windows and take a right into the recreation room and stop before you reach the pool table.}%
          \label{fig:r2r-pool}%
    \end{subfigure}
    \hfill
    \begin{subfigure}[b]{0.47\textwidth}
        \centering
    \hspace*{-2cm}\includegraphics[width=\textwidth]{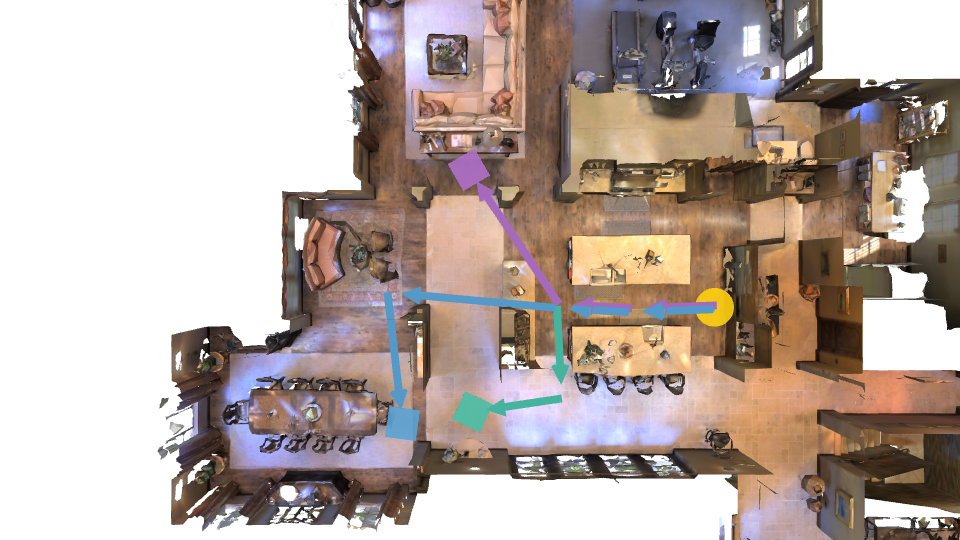} 
    \caption{\textbf{R2R \success{}:} Go between the counters, turn left, turn right, and stop before the display and dining room.}%
    \label{fig:r2r-image}%
    \end{subfigure}
    \hfill
    \begin{subfigure}[b]{0.47\textwidth}
        \centering
           \includegraphics[width=\textwidth]{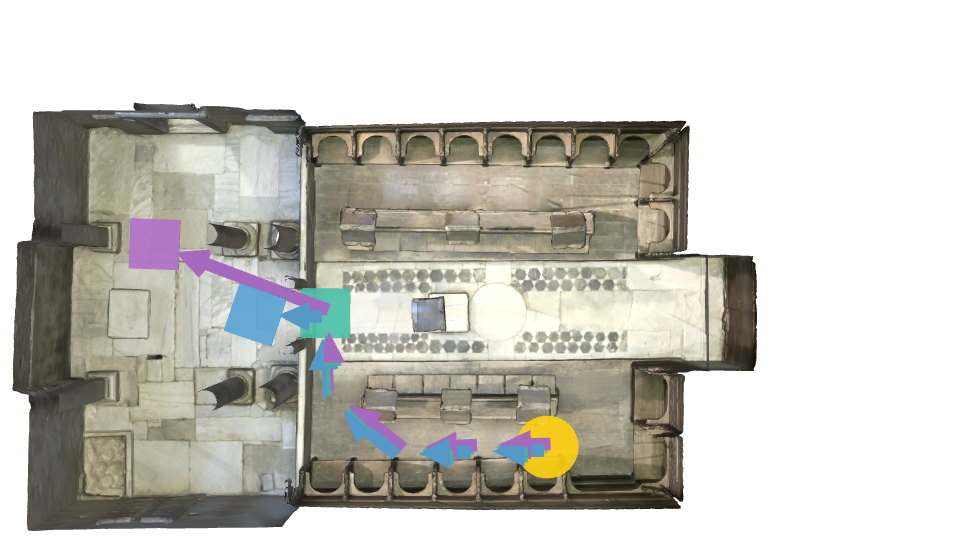}\hspace*{-3cm}
    \caption{\textbf{R2R \success{}:} Turn right and head towards the end. Once you reach the end make a right and stop.}%
    \label{fig:r2r-church}%
    \end{subfigure}
   \hfill
    \begin{subfigure}[b]{0.47\textwidth}
      \centering
    \includegraphics[width=0.8\textwidth]{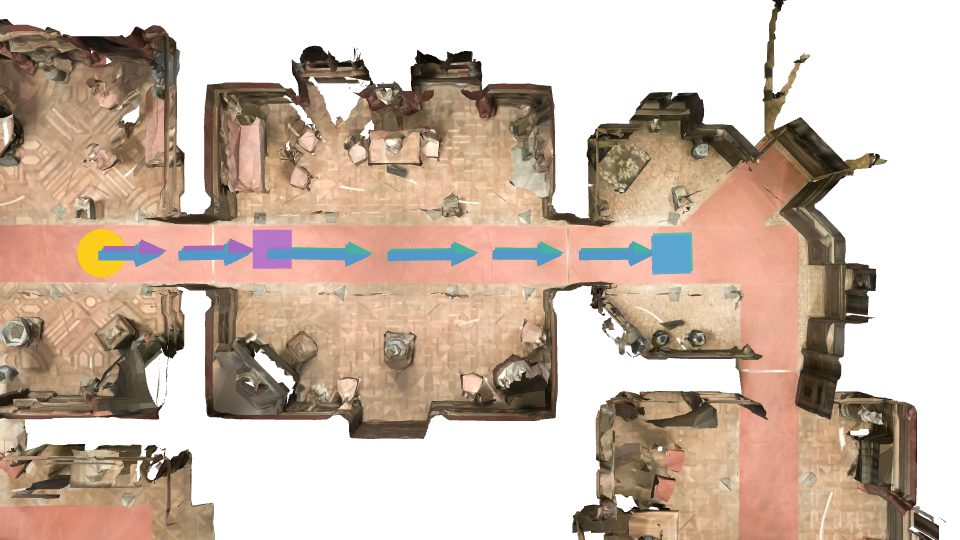} 
    \caption{\textbf{R2R \success{}:} Walk straight out the door in front of you and follow the red carpet. Keep going through the room with the ropes and stop when you enter the next room with ropes.}%
    \label{fig:r2r-redcastle4}%
    \end{subfigure}
    
    \caption{When navigating in new houses, our \airbert~model not only successfully recognizes the closet room in (a) and (b), pool table (c), living room (d), but also generalizes better to challenging environments, such as the church (e) and castle (f).}
    \label{fig:r2r_new_houses}
\end{figure*}

\p{New houses.}
In Figure~\ref{fig:r2r_new_houses}, we compare predicted paths from \vlnbert~and \airbert~in new houses beyond the training environments.
Benefiting from \airbnb~dataset that provides diverse visual environments in pretraining, our \airbert~model generalizes better to recognize different room types in new houses (see Figure~\ref{fig:r2r-closet}-\ref{fig:r2r-image}), and performs better on significantly different environments such as a church (Figure~\ref{fig:r2r-church}) or castle (Figure~\ref{fig:r2r-redcastle4}).

\begin{figure*}[t]
    \centering
    \begin{subfigure}[b]{0.47\textwidth}
    \centering
    \includegraphics[width=\textwidth]{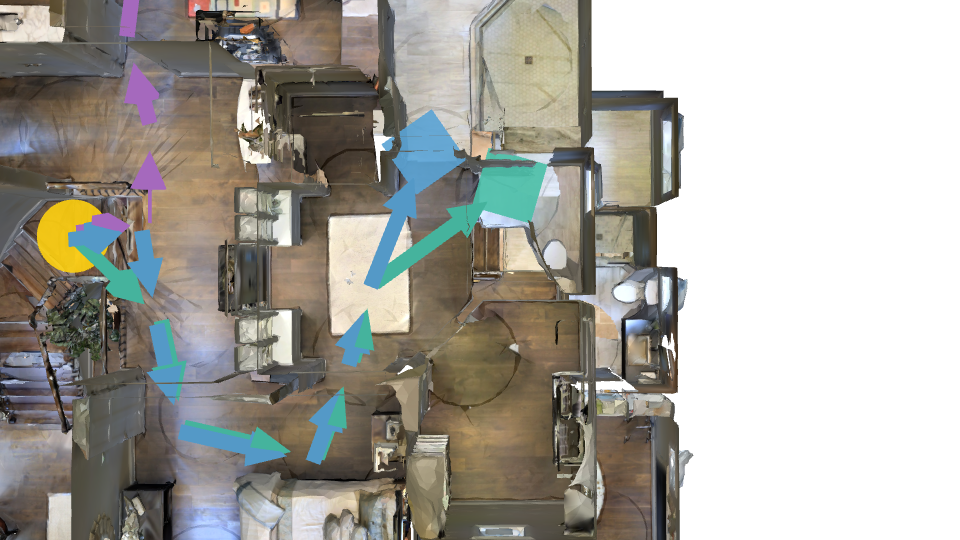} 
    \hspace*{-3cm}\caption{\textbf{R2R \success{}:} Walk up the stairs and take a right. Walk into the bedroom and take a left . Take another left at the {\color{red} night stand} and walk out of the bedroom. Wait by the toilet in the second door on the right.}%
    \label{fig:r2r-word}%
    \end{subfigure}
    \hfill
    \begin{subfigure}[b]{0.47\textwidth}
    \centering
    \includegraphics[width=\textwidth]{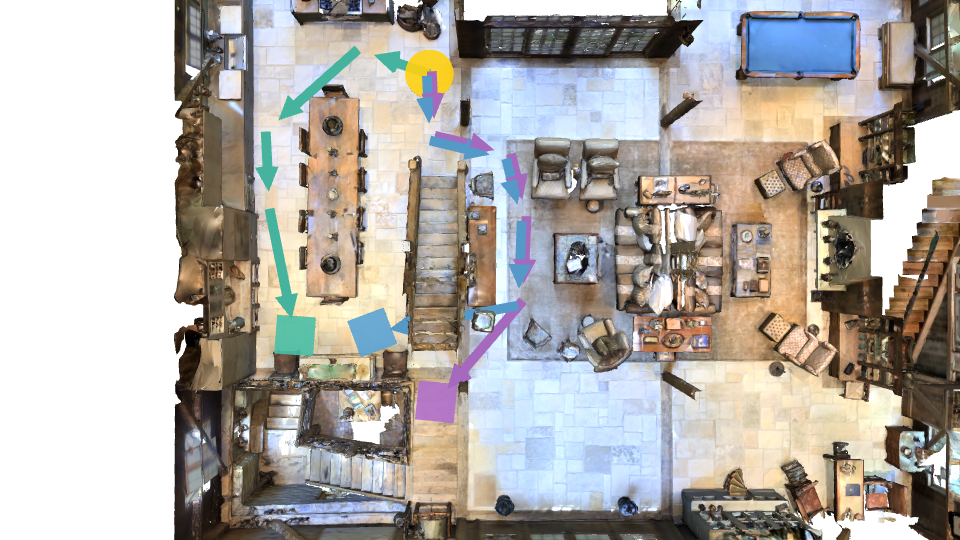} 
    \caption{\textbf{R2R \success{}:} Go straight past the table and chairs then turn left and continue to go past the table and chairs. Wait near the white {\color{red} antique furniture} with the two chairs on on each side.}%
    \label{fig:r2r-word4}%
    \end{subfigure}
    
    \begin{subfigure}[b]{0.47\textwidth}%
    \centering
    \includegraphics[width=\columnwidth]{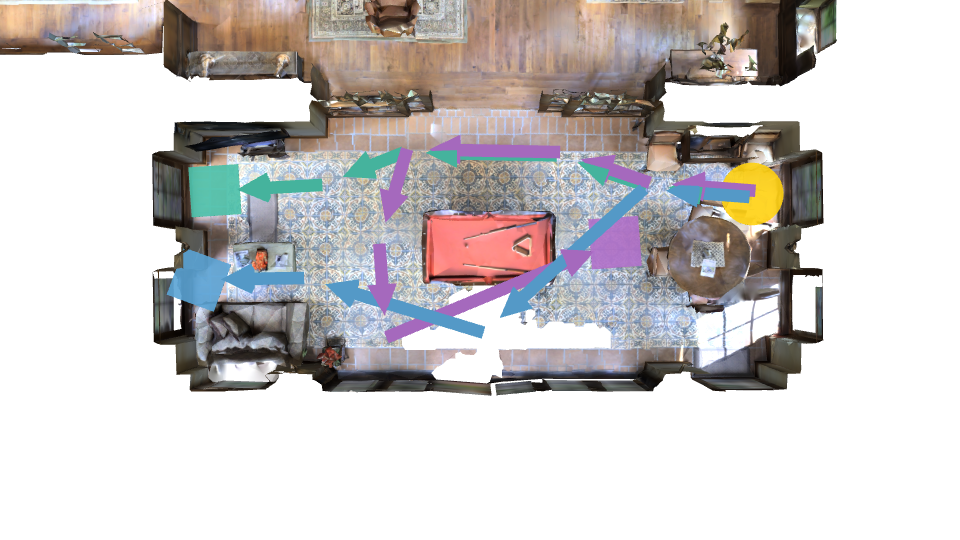} 
    \caption{\textbf{REVERIE \success{}:} Walk past the  {\color{red} pool table} and towards the TV on the far side of the room and grab the coffee table that is located in front of the couch}%
    \end{subfigure}  
    \hfill
    \begin{subfigure}[b]{0.47\textwidth}%
    \centering
    \includegraphics[width=\columnwidth]{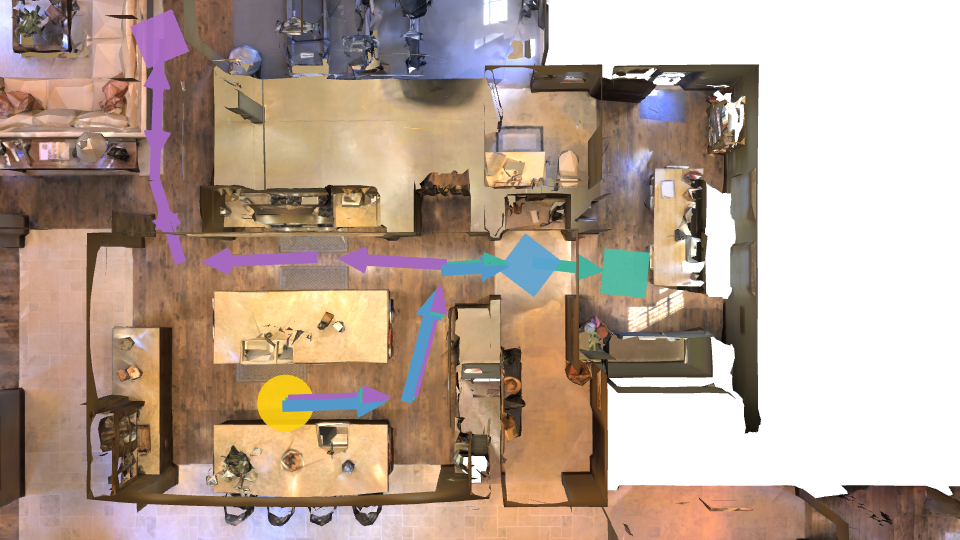} 
    \caption{\textbf{REVERIE \success{}:} Please go to the pantry room with the two large freezers and kitchen appliances on the large table and reset the flipped breaker in the breaker panel box to the right of the  {\color{red} freezers}}%
    \end{subfigure}
    \caption{The \airbert~model outperforms \vlnbert~to recognize rare or even unseen objects in training set. (a) Rare object ``\emph{night stand}''; (b) unseen object ``\emph{antique furniture}''; (c) rare object ``\emph{pool table}''; and (d) unseen object ``\emph{freezer}''.}
    \label{fig:noun-phrases}
\end{figure*}

\p{New objects.}
\airbert~also improves the understanding of new objects in home environments, \eg~through noun phrases related to household objects.
As shown in Figure~\ref{fig:noun-phrases}, it is successful at following instructions containing noun phrases that rarely occur or are even unseen on the training set, while the \vlnbert~model that is trained on a large image-caption corpus not pertaining to houses fails.

\begin{figure*}[t]
    \centering
    \begin{subfigure}[b]{0.47\textwidth}
    \centering
    \includegraphics[width=\columnwidth]{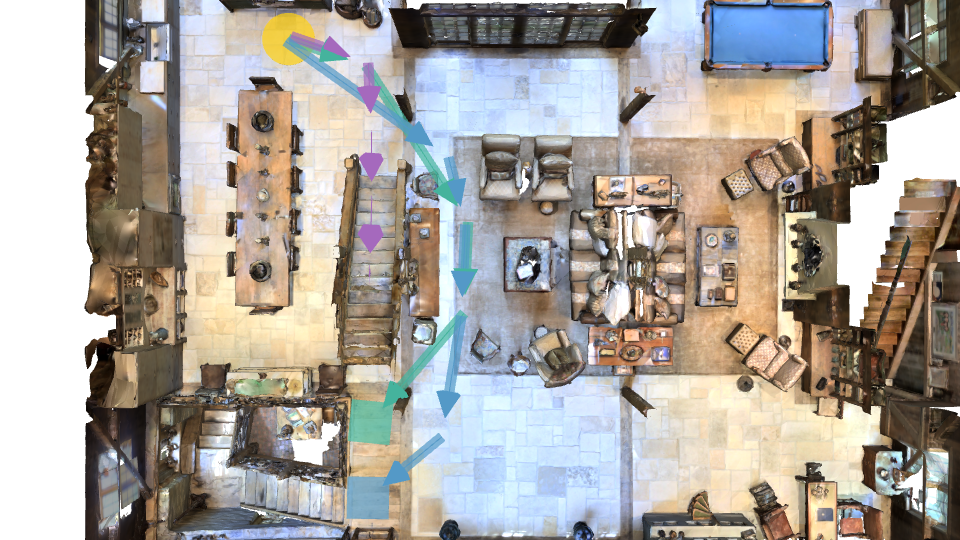}%
    \caption{\textbf{R2R \success{}:} Walk from dining room to living room turning slightly right before lounge chairs, walk straight following cabinet. Turn slight right and stop at stairs.}%
    \label{fig:r2r-shuff1}%
\end{subfigure}
\hfill
\begin{subfigure}[b]{0.47\textwidth}
    \centering
    \includegraphics[width=\columnwidth]{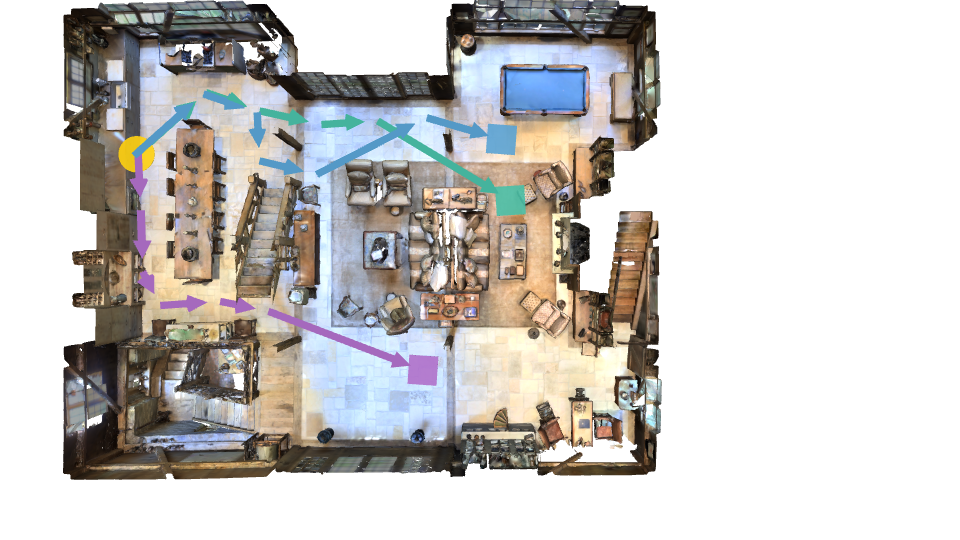} 
    \caption{\textbf{R2R \success{}:} Walk on into the kitchen and turn to the right. Walk past the staircase, behind the chairs. Walk to the right of the pillar. Stop and wait by the footstool.}%
    \label{fig:r2r-shuff3}%
\end{subfigure}

\begin{subfigure}[b]{0.47\textwidth}
    \centering
    \includegraphics[width=\columnwidth]{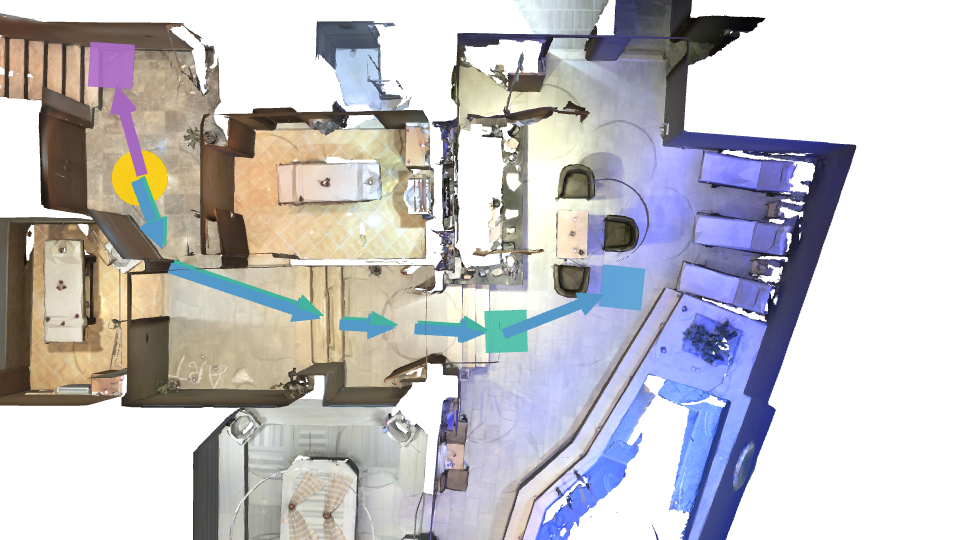} 
    \caption{\textbf{R2R \success{}:} Walk out of the hallway and turn left. Walk down the steps and through the wooden doors. Walk down the steps and stop.}%
    \label{fig:r2r-shuff6}%
\end{subfigure}
%
\hfill
\begin{subfigure}[b]{0.47\textwidth}
    \centering
    \includegraphics[width=\columnwidth]{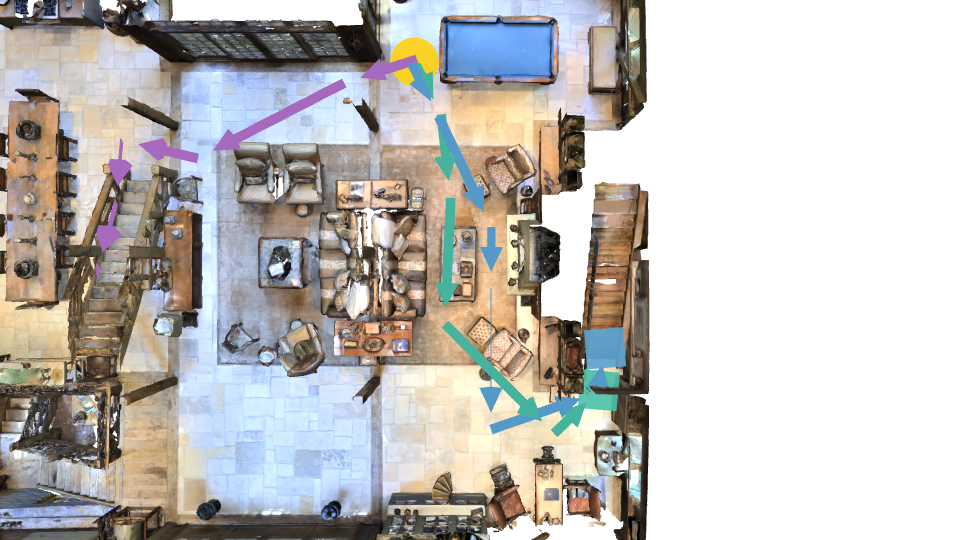} 
    \caption{\textbf{R2R \success{}:} Go straight passed the coffee table turn left and go through the left door to the stairs. Stop in front of the stairs.}%
    \label{fig:r2r-shuff10}%
\end{subfigure} 

\caption{Examples in similar environments and instructions to the training set. The improvements of \airbert~model can be contributed to the shuffling loss in pretraining.}
\label{fig:shuffling}
\end{figure*}

\p{Similar environments and instructions.}
Figure~\ref{fig:shuffling} displays examples where the environments and the instructions are similar to those on the training set, with the aim to show that the shuffling loss in pretraining also benefits learning.
For example, in Figure~\ref{fig:r2r-shuff1},  the VLN-BERT agent \pfinal{} focuses on the stairs (in the last step) and goes upstairs incorrectly, whereas \airbert~learns to consider intermediate steps such as ``\emph{lounge chairs}'' and ``\emph{cabinet}'' besides the last step by learning from the shuffling task.
Similarly, in Figure~\ref{fig:r2r-shuff6}, we see that the \vlnbert~agent stops at the wrong stairs, while \airbert~considers intermediate steps such as ``\emph{hallway}'' and ``\emph{wooden doors}'', and ends within the acceptable range of 3m from the goal.

%

\begin{figure*}[t]
\centering 
\begin{subfigure}[b]{0.47\textwidth}
    \centering
    \includegraphics[width=\columnwidth]{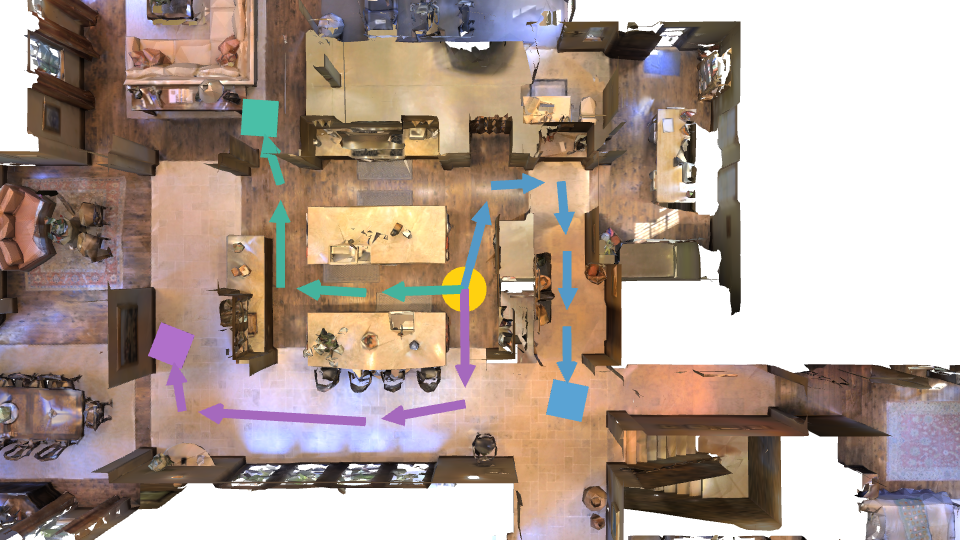} 
    \caption{\textbf{R2R \failure{}:} Walk between the two kitchen islands and then turn right. Pass through the stone archway and stop just after you pass through it. Wait there.}%
    \label{fig:r2r-failure1}%
\end{subfigure}
\hfill
\begin{subfigure}[b]{0.47\textwidth}%
    \centering
    \includegraphics[width=\columnwidth]{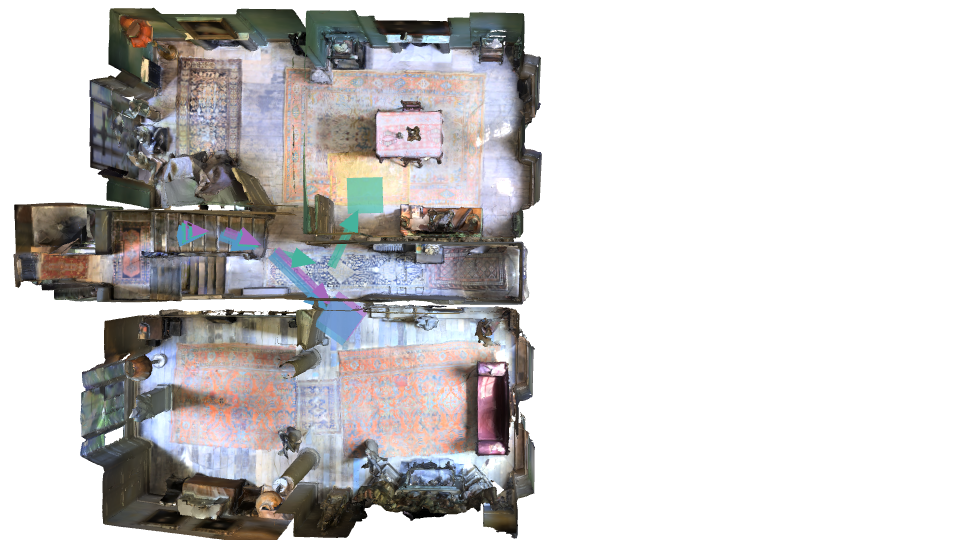} 
    \caption{\textbf{R2R \failure{}:} Exit the bathroom and go down the stairs. Enter the last doorway on the left and stop just before stepping on the rug.}%
    \label{fig:r2r-failure2}%
\end{subfigure}

\begin{subfigure}[b]{0.47\textwidth}%
    \centering
    \includegraphics[width=\columnwidth]{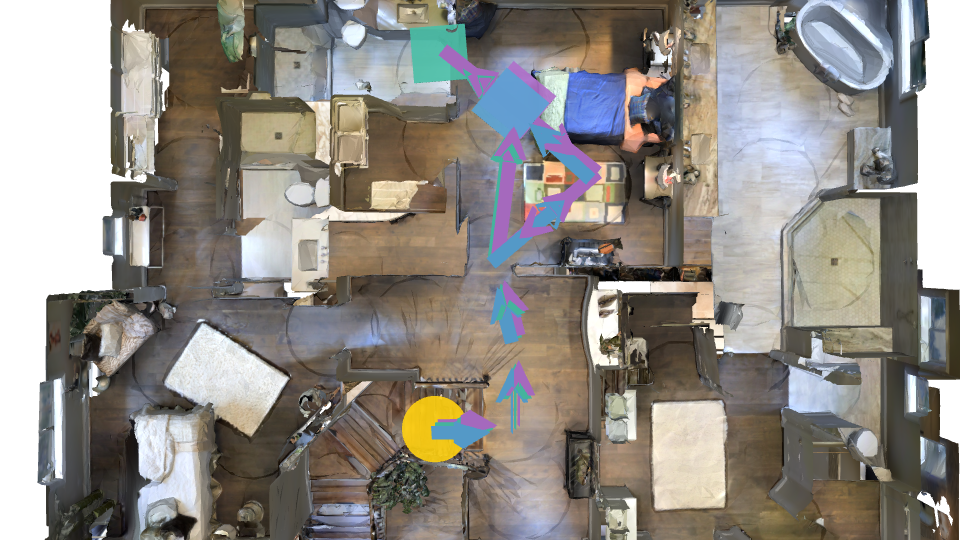} 
    \caption{\textbf{REVERIE \failure{}:} go to level 3 bathroom in the first bedroom left of the stairs and grab the mirror on the wall}%
    \label{fig:reverie-failure1}%
\end{subfigure} 
%
\hfill
\begin{subfigure}[b]{0.47\textwidth}%
    \centering
    \includegraphics[width=\columnwidth]{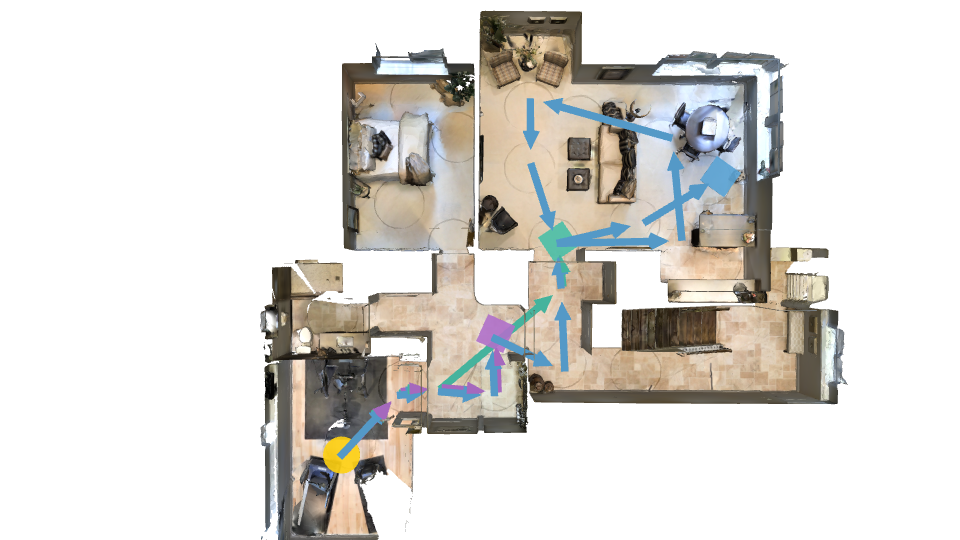} 
    \caption{\textbf{REVERIE \failure{}:} Go to the lounge on this level and polish the black leather armchair in the corner}%
    \label{fig:reverie-failure2}%
\end{subfigure} 
\caption{Failure cases for both \vlnbert~and \airbert~models.}
\label{fig:failures}
\end{figure*}

\p{Failure cases.}
Figure~\ref{fig:failures} presents some failure cases for both \vlnbert{} and \airbert{}.
It reveals that current models still struggle to deal with relationships such as ``\emph{between}'' (Figure~\ref{fig:r2r-failure1}), or directional instructions such as ``\emph{on the left}'' (Figure~\ref{fig:r2r-failure2}).
Similar failures are also highlighted by Table 5 of the main paper where we show that models fail to choose the correct instruction when a direction keyword (left/right) is switched.

\subsection{Few-shot Learning on VLN}

As mentioned in the main paper, we present complete results for the few-shot learning evaluation, along with standard deviations in Table~\ref{tab:fsl-complete}.
While the performance on the seen validation houses fluctuates a lot (also due to changing the environment in the seen validation set), unseen validation is very stable.
Recall that \vlnbert~achieves an unseen validation performance of 27\% and 37\% with 1 and 6 training environments respectively.
On the other hand, \airbert~achieves a superior 49.5\% and 58\% -- an absolute improvement of $\sim$22\% in both cases.

\input{tables/fsl-complete}

%% file: tables/r2r_generative.tex
\begin{table*}
	\small
	\centering
	\begin{tabular}{l|cccc|cccc|cccc} \toprule
		\multirow{2}{*}{Methods} & \multicolumn{4}{c|}{Validation Seen} & \multicolumn{4}{c|}{Validation Unseen} & \multicolumn{4}{c}{Test Unseen} \\
		& TL & NE & SR & SPL & TL & NE & SR & SPL & TL & NE & SR & SPL \\ \midrule
		Seq2Seq-SF \cite{anderson2018r2r} & 11.33 & 6.01 & 39 & - & 8.39 & 7.81 & 22 & - & 8.13 & 7.85 & 20 & 18 \\
		Speaker-Follower \cite{fried2018speaker} & - & 3.36 & 66 & - & - & 6.62 & 35 & - & 14.82 & 6.62 & 35 & 28 \\
		PRESS \cite{li2019press} & 10.57 & 4.39 & 58 & 55 & 10.36 & 5.28 & 49 & 45 & 10.77 & 5.49 & 49 & 45 \\
		EnvDrop \cite{tan2019envdrop} & 11.00 & 3.99 & 62 & 59 & 10.70 & 5.22 & 52 & 48 & 11.66 & 5.23 & 51 & 47 \\
		PREVALENT \cite{hao2020prevalent} & 10.32 & 3.67 & 69 & 65 & 10.19 & 4.71 & 58 & 53 & 10.51 & 5.30 & 54 & 51 \\
		Rec (no init. OSCAR) \cite{hong2021recurrentvln} & 9.78 & 3.92 & 62 & 59 & 10.31 & 5.10 & 50 & 46 & 11.15 & 5.45 & 51 & 47 \\
		Rec (OSCAR) \cite{hong2021recurrentvln} & 10.79 & 3.11 & 71 & 67 & 11.86 & 4.29 & 59 & 53 & 12.34 & 4.59 & 57 & 53 \\
		Rec (PREVALENT) \cite{hong2021recurrentvln} & 11.13 & 2.90 & 72 & 68 & 12.01 & 3.93 & 63 & 57 & 12.35 & 4.09 & 63 & 57 \\ \midrule
		Rec (\vilbert)  & 11.16 & 2.54 & 75 & 71 & 12.44 & 4.20 & 60 & 54 & - & - & - & - \\
		Rec (\vlnbert) & 10.95 & 3.37 & 68 & 64 & 11.33 & 4.19 & 60 & 55 & - & - & - & - \\
		Rec (\airbert) & 11.09 & 2.68 & 75 & 70 & 11.78 & 4.01 & 62 & 56 & 12.41 & 4.13  & 62 & 57 \\  \bottomrule
	\end{tabular}
	\caption{Navigation performance of different generative models on the R2R dataset.}
	\label{tab:r2r_generative_results}
\end{table*}

%% file: tables/fsl-complete.tex
\begin{table*}[t]
\centering
\tabcolsep=0.14cm
\begin{tabular}{@{\extracolsep{1mm}}cc ccccc ccccc@{}}
\toprule
\multirow{2}{*}{\# Env.} &
\multirow{2}{*}{Traj.} & 
\multicolumn{5}{c}{Val Seen~SR} &
\multicolumn{5}{c}{Val Unseen~SR}
\\ 
\cline{3-7} \cline{8-12}
& & PL & NE & SPL & OSR & SR & PL & NE & SPL & OSR & SR \\ 
\midrule
1  & Rand 
    & 10.97	& 5.36 & 0.44 & 63.74 &	47.87 $\pm 0.03$
    & 10.84	& 4.86 & 0.51 & 68.46 & 54.48 $\pm 0.04$ \\
\midrule
6  & Rand
    & 9.84 & 5.49 & 0.47 & 65.93 & 50.00 $\pm 0.02$
    & 9.55 & 4.55 & 0.55 & 70.89 & 57.97 $\pm 0.01$ \\
\midrule
61 & Rand
    & 10.91 & 4.87 & 0.60 & 76.23 & 64.24
    & 9.50  & 3.70 & 0.62 & 76.24 & 65.60  \\
61 & \cite{tan2019envdrop}
    & 10.59 & 3.21 & 0.69 & 80.71 & 73.85 
    & 10.03 & 3.24 & 0.63 & 78.45 & 68.67 \\
\bottomrule
\end{tabular}

\caption{Performance of Airbert on R2R few-shot evaluation.
During training, only a subset of the Matterport~\cite{Matterport3D} environments are accessible.}
\label{tab:fsl-complete}
\end{table*}